\documentclass[]{template}

\usepackage[utf8]{inputenc}             
\usepackage[T1]{fontenc}                
\usepackage{url}                        
\usepackage{booktabs}                   
\usepackage{multirow}
\usepackage{colortbl}
\usepackage{multicol}
\usepackage{amsfonts}                   
\usepackage{nicefrac}                   
\usepackage{microtype}                  
\usepackage[dvipsnames]{xcolor}         

\usepackage{latexsym}

\usepackage{graphicx}
\usepackage{float}
\usepackage{subcaption}
\usepackage{wrapfig}
\usepackage{lipsum}

\usepackage{bm}

\usepackage{tabularx} 
\usepackage{ragged2e} 
\newcolumntype{L}{>{\RaggedRight\hangafter=1\hangindent=0em}X}

\usepackage{enumitem}

\usepackage{amsmath}
\usepackage{amssymb}
\usepackage{mathtools}
\usepackage{amsthm}
\usepackage{algorithm}
\usepackage{algpseudocode}
\usepackage{float}

\setboolean{logo}{true}    

\usepackage[capitalize,noabbrev]{cleveref}
\crefname{section}{§}{§§}
\Crefname{section}{§}{§§}

\usepackage{calligra}
\DeclareMathAlphabet{\mathcalligra}{T1}{calligra}{m}{n}

\usepackage{pifont}

\theoremstyle{plain}

\theoremstyle{definition}

\theoremstyle{remark}

\renewcommand{\paragraph}[1]{\vspace{1mm}\noindent\textbf{#1}}

\DeclareCaptionLabelFormat{cont}{#1~#2\alph{ContinuedFloat}}
\usepackage[most]{tcolorbox}
\tcbset{
  promptbox/.style={
    top=10pt,
    colback=lightgray!20,
    colframe=Black,
    colbacktitle=NavyBlue,
    enhanced,
    center,
    attach boxed title to top center={yshift=-0.1in,xshift=0.0in},
    boxed title style={boxrule=0pt,colframe=white,},
  }
}
\newtcolorbox{promptbox}[2][]{promptbox, title=#2,#1}
\tcbset{
  takeawaybox/.style={
    top=10pt,
    colback=lightgray!20,
    colframe=Black,
    colbacktitle=BurntOrange,
    enhanced,
    center,
    attach boxed title to top center={yshift=-0.1in,xshift=0.0in},
    boxed title style={boxrule=0pt,colframe=white,},
  }
}
\newtcolorbox{takeawaybox}[2][]{takeawaybox, title=#2,#1}
\tcbset{
  observationbox/.style={
    top=10pt,
    colback=lightgray!20,
    colframe=Black,
    colbacktitle=YellowGreen,
    enhanced,
    center,
    attach boxed title to top center={yshift=-0.1in,xshift=0.0in},
    boxed title style={boxrule=0pt,colframe=white,},
  }
}
\newtcolorbox{observationbox}[2][]{observationbox, title=#2,#1}

\usepackage{xspace}

\newcommand\blfootnote[1]{%
  \begingroup
  \renewcommand\thefootnote{}\footnote{#1}%
  \addtocounter{footnote}{-1}%
  \endgroup
}

\newcommand{\methodname}{EndoCoT\xspace}
\definecolor{failred}{HTML}{B03A2E}
\definecolor{botblue}{HTML}{0056B3}

\usepackage{CJK}

\title{\methodname: Scaling Endogenous Chain-of-Thought Reasoning in Diffusion Models}

\author[1,2]{Xuanlang Dai}
\author[1,3]{Yujie Zhou}
\author[1,4]{Long Xing}
\author[1,3]{Jiazi Bu}
\author[1,5]{Xilin Wei}
\author[1,3]{Yuhong Liu}
\author[1,6]{Beichen Zhang}
\author[1]{$\dagger$Kai Chen}
\author[1]{$\dagger$Yuhang Zang}

\affil[1]{Shanghai AI Laboratory}
\affil[2]{Xi'an Jiaotong University}
\affil[3]{Shanghai Jiaotong University}
\affil[4]{University of Science and Technology of China}
\affil[5]{Fudan University}
\affil[6]{The Chinese University of Hong Kong}

\begin{abstract}

Recently, Multimodal Large Language Models (MLLMs) have been widely integrated into diffusion frameworks primarily as text encoders to tackle complex tasks such as spatial reasoning. 
However, this paradigm suffers from two critical limitations: 
(i) \textbf{MLLMs text encoder exhibit insufficient reasoning depth}. 
Single-step encoding fails to activate the Chain-of-Thought process, which is essential for MLLMs to provide accurate guidance for complex tasks.
(ii) \textbf{The guidance remains invariant during the decoding process}.
Invariant guidance during decoding prevents DiT from progressively decomposing 
complex instructions into actionable denoising steps, even with correct MLLM encodings. 
To this end, we propose Endogenous Chain-of-Thought (EndoCoT), 
a novel framework that first activates MLLMs' reasoning potential by
iteratively refining latent thought states through an iterative thought guidance module, 
and then bridges these states to the DiT's denoising process.
Second, a terminal thought grounding module is applied to ensure the reasoning trajectory
remains grounded in textual supervision by aligning the final state with ground-truth answers.
With these two components, the MLLM text encoder delivers meticulously reasoned guidance, 
enabling the DiT to execute it progressively and ultimately solve complex tasks in a step-by-step manner.
Extensive evaluations across diverse benchmarks (e.g., Maze, TSP, VSP, and Sudoku)
achieve an average accuracy of 92.1\%, outperforming the strongest baseline by 8.3 percentage points. 

\end{abstract}

\begin{document}

\blfootnote{$\dagger$ Corresponding authors: Kai Chen ( chenkai@pjlab.org.cn@pjlab.org.cn), Yuhang Zang (zangyuhang@pjlab.org.cn)}
\blfootnote{$*$ Code is at \url{https://github.com/InternLM/EndoCoT}}

\maketitle

\section{Introduction}\label{sec:intro}

\begin{figure}[t]
        \centering
        \includegraphics[width=1\linewidth]{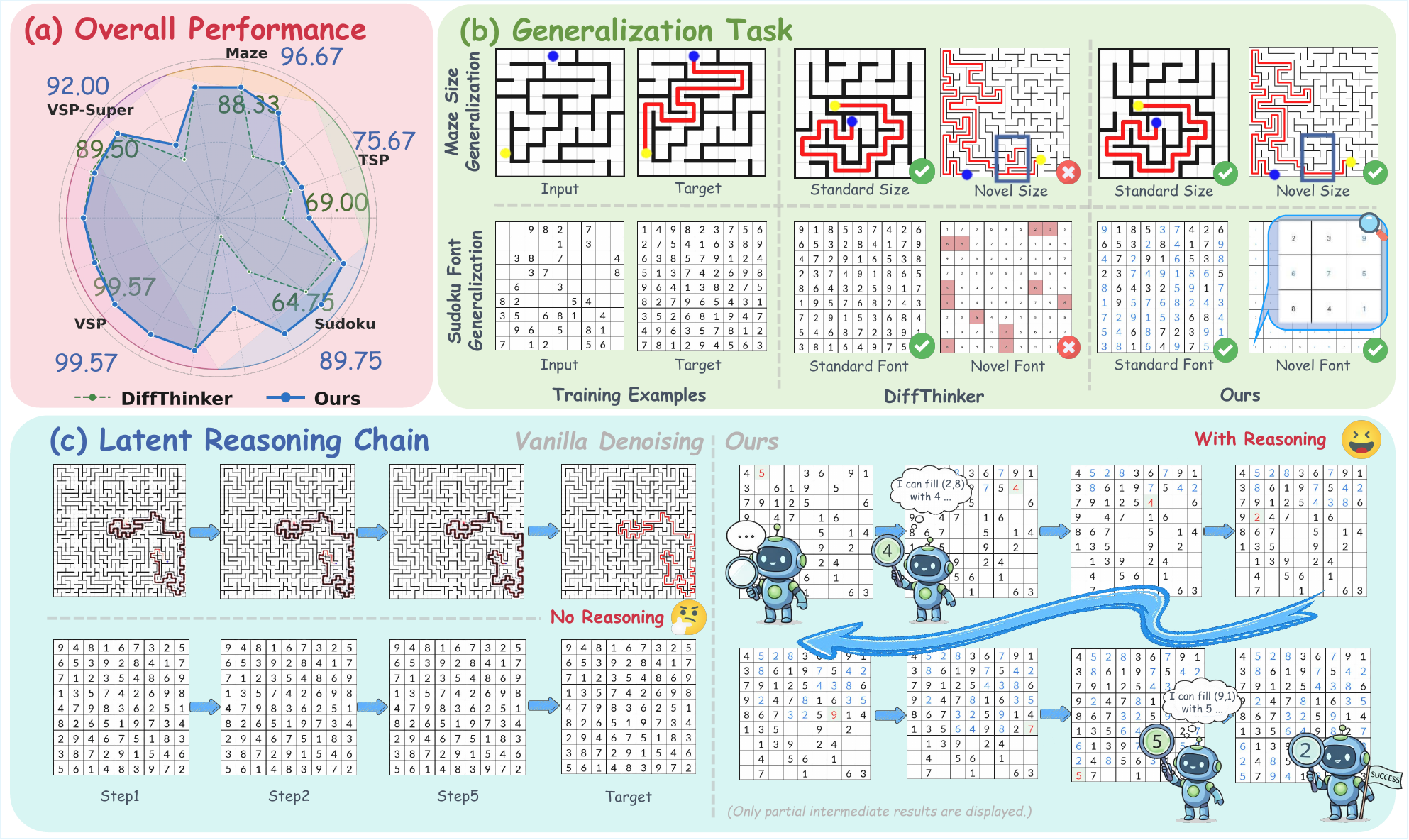}
        \vspace{-12pt}
        \caption{\textbf{\methodname enables endogenous chain-of-thought reasoning.}
        \textbf{(a)} Radar plot showing \methodname outperforms baselines across all benchmarks.
        \textbf{(b)} On visual reasoning tasks requiring generalization (maze size, Sudoku font), previous work~\cite{he2025diffthinker} fails on novel domains while \methodname consistently generalizes correctly.
        \textbf{(c)} Vanilla denoising (left) commits to solutions early without reasoning, while our approach (right) enables interpretable, step-by-step reasoning chains.}
        \label{fig:teaser}
        \vspace{-12pt}
\end{figure}

Diffusion models~\cite{lipman2022flow, ho2020denoising, song2020denoising, song2020score} have revolutionized visual generation, achieving unprecedented levels of photorealism and diversity.
Recent efforts integrate Multimodal Large Language Models (MLLMs) \cite{bai2025qwen3vltechnicalreport, liu2026ministral, wu2025qwen, blackforestlabs2025flux2} as text encoders to augment semantic understanding and logical alignment.
Yet despite these advances, they struggle with tasks requiring \textit{step-by-step logical reasoning}, such as solving mazes, planning Traveling Salesperson Problem (TSP) routes, or completing Sudoku puzzles, where solutions must adhere to strict sequential constraints and generalize to novel configurations.

Current paradigms relegate MLLMs to the role of \textit{static conditional encoders}, computing text embeddings once at the beginning of generation.
This passive integration fails to exploit the dynamic reasoning potential of MLLMs.
Even recent explicit attempts like DiffThinker~\cite{he2025diffthinker} to ``inject'' reasoning into diffusion models result in superficial alignment rather than genuine cognitive processing.
While previous work~\cite{he2025diffthinker} reports improved metrics on specific benchmarks, they produce fragile solutions that fail catastrophically when generalized to novel domains, see Fig.~\ref{fig:teaser}(b).
Our analysis reveals why these failures occur: despite being built on powerful MLLMs, these models do not actually perform reasoning \textit{during} generation.
Instead, as shown in in Fig.~\ref{fig:teaser}(c), they commit to their final solution within the first few denoising steps and merely refine visual quality thereafter.
This suggests that without an \textit{endogenous} mechanism, forced alignment results in fragile pattern matching rather than robust, iterative problem-solving.

Motivated by these persistent failures in supervised fitting, we conduct a systematic empirical analysis (Sec.~\ref{sec:analysis}) that identifies two key bottlenecks: (1) \textit{limited single-step reasoning}: MLLMs cannot encode all necessary logical constraints in a single forward pass; (2) \textit{static-guidance failure}: Diffusion Transformers (DiTs) cannot maintain alignment with complex logical constraints when conditioned on static embeddings.
Chain-of-Thought (CoT) reasoning~\cite{wei2022chain} has shown how to overcome similar limitations in MLLMs~\cite{yang2025qwen3, anthropic2026claudeopus46} through iterative refinement, but diffusion models currently lack an analogous mechanism.

To this end, we propose \textit{Endogenous} CoT (\textbf{\methodname}), a framework that enables diffusion models to perform self-guided reasoning by iteratively exploring the semantic latent space.
Specifically, \methodname encompasses two primary components: 
(1) \textbf{Iterative Thought Guidance}: iteratively updates latent states in the MLLM to create a genuine CoT-like reasoning process
and establishes correspondence with DiT’s denoising process.
(2) \textbf{Terminal Thought Grounding}: aligns the MLLM's final reasoning state with ground-truth answers,
ensuring the reasoning trajectory remains grounded in textual supervision and preventing cumulative drift in the terminal output.
Consequently, \methodname serves as a complete end-to-end pipeline, seamlessly integrating reasoning and generation within a unified diffusion framework.
Experiments are conducted across four diverse reasoning tasks: Maze, TSP, Visual Spatial Planning (VSP), and Sudoku,
demonstrating consistently superior performance over state-of-the-art baselines including DiffThinker~\cite{he2025diffthinker} and Qwen3-VL-8B~\cite{bai2025qwen3vltechnicalreport}.
As summarized in Fig.~\ref{fig:teaser}(a), \methodname outperforms baselines across all benchmarks.
Moreover, as task complexity scales, \methodname maintains exceptional accuracy: it achieves 90\% on Maze-32 and 95\% on Sudoku-35, outperforming the strongest baseline by 25\% and 40\% respectively.
Unlike vanilla baselines that commit early and fail catastrophically, \methodname exhibits interpretable, step-by-step reasoning chains (see Fig.~\ref{fig:teaser}(c)). 
Our main contributions are summarized as follows:

\begin{enumerate}
    \item To the best of our knowledge, we present the first diffusion framework that enables genuine chain-of-thought reasoning via iterative latent state refinement, rather than pre-computing solutions in a single pass.
    \item Through comprehensive layer-wise sensitivity and attention entropy analysis, we localize where reasoning arises in diffusion models and identify the two key bottlenecks that limit prior methods (Sec.~\ref{sec:analysis}).
    \item \methodname achieves 25-40\% gains over prior work on complex visual reasoning benchmarks, enables controllable inference-time scaling, and yields clearer and more controllable transformation trajectories in image editing tasks.
\end{enumerate}
\section{Related Work}\label{sec:related_work}
\vspace{0.5em}

\noindent \textbf{Reasoning in Multimodal Large Language Models}
Chain-of-thought (CoT) and other test-time scaling strategies \cite{zhang2025survey, yao2023tree, besta2024graph, zheng2025monte} have proven effective in autoregressive large language models (LLMs). Recent works \cite{yan2025videochat, agarwal2025art} extend this paradigm to multimodal settings. OpenAI \cite{openai2025thinkingwithimages} introduces “Think with Images”. Subsequent works \cite{rotstein2025pathways, zhang2025thinking, tong2025thinking, qin2025chain, wang2025video, fei2024video} further extend this line of research to “Thinking with Video”, using visual content as external evidence to support multi-step reasoning. Latent Sketchpad \cite{zhang2025latent} proposes an interleaved autoregressive generation of text and visual latents. However, enabling reasoning in diffusion models (DMs) remains challenging. Since DMs typically rely on fixed-length text encoders \cite{radford2021learning, raffel2020exploring}, their capacity to incorporate multi-step reasoning into generation is limited. Recent works \cite{wu2025qwen} introduce vision-language models (VLMs) as encoders, providing richer semantic representations that facilitate reasoning-conditioned generation in diffusion models.
\vspace{0.5em}

\noindent \textbf{Reasoning in Diffusion Model}
Several works \cite{jiao2025thinkgen, kou2026think} explore incorporating reasoning signals into DMs by injecting textual reasoning traces into the conditioning inputs. While those approaches enable reasoning-aware generation, the MMDiT is often treated as a conditional decoder. This over-reliance on VLMs leads to a decoupled pipeline in which the MLLM mainly acts as a prompt enhancer. Recently, approaches \cite{zhang2025image,wu2025chronoedit,wang2026bigvideoreasoningsuite} have leveraged video priors to perform complex editing, essentially treating logical state transitions as temporal sequences. However, these methods rely on the inherent smoothness of video models to ``reason'' through changes, which may not translate to discrete logical tasks. DiffThinker \cite{he2025diffthinker} takes a step toward internalizing reasoning by exploring the reasoning potential of MMDiT directly. Works like \cite{gao2025d,gu2024dart} have attempted visual generation under the next-token autoregressive paradigm.
Despite these advancements, existing frameworks still cannot perform genuine, iterative chain-of-thought reasoning within the diffusion process itself.
In this work, we enable endogenous CoT reasoning through iterative latent state refinement, achieving robust logical planning and spatial grounding.
\vspace{0.5em}

\noindent \textbf{Latent Reasoning}
Latent reasoning has been validated in text-only domains, enabling multi-step reasoning in continuous latent spaces \cite{hao2024training, zhang2025soft, tan2025think}. Latent-space reasoning compresses long reasoning chains and supports tree-structured exploration, improving inference efficiency and diversity \cite{shen2025codi, zheng2026imitationreinforcementlearningactive}. Motivated by this line of work, we train diffusion models end-to-end with latent tokens to enable test-time scaling within the diffusion process.

\section{Analysis of Reasoning in Diffusion Models}
\label{sec:analysis}
To understand why current diffusion models fail at complex visual reasoning despite integrating powerful MLLMs, we first conduct a systematic empirical analysis.
These analyses directly motivate the design of our \methodname framework and reveal fundamental bottlenecks that limit performance. All experiments are conducted on  Qwen-Image-Edit-2511 \cite{wu2025qwen}.

\begin{figure}[t]
        \centering
        \includegraphics[width=1\linewidth]{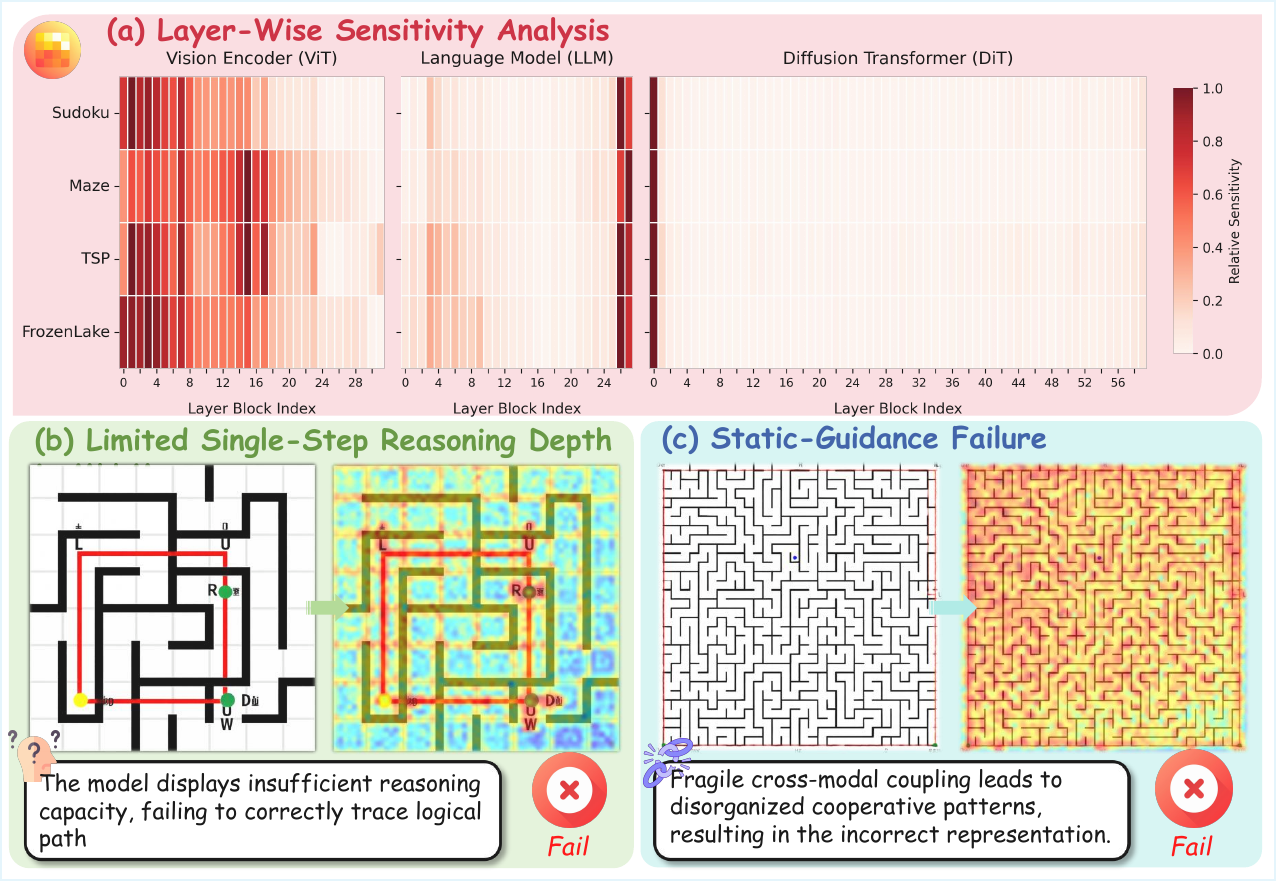}
        \vspace{-12pt}
        \caption{
        \textbf{(a) Layer-wise sensitivity} across Vision Encoder, LLM, and DiT components (red: high sensitivity, white: low sensitivity).
        \textbf{(b) Limited single-step reasoning}: DiT performs spatial grounding but trajectory violates constraints.
        \textbf{(c) Static-guidance failure}: Dense topologies cause attention entropy to become diffuse.
        }
        \label{fig:analysis/fail}
        \vspace{-12pt}
\end{figure}

\noindent \textbf{Layer-Wise Sensitivity Analysis.}
To investigate the internal dynamics of how reasoning capabilities emerge, we conduct a layer-wise sensitivity analysis on a representative image editing model. Specifically, we quantify the relative sensitivity by measuring the magnitude of activation responses across the distinct layer blocks of the model's three core components: the Vision Encoder, the Language Model, and the Diffusion Transformer.

As shown in Fig. \ref{fig:analysis/fail}(a), peak sensitivity is highly concentrated within the Vision Encoder and precisely at the architectural junction between the LLM (terminal layers) and the DiT (initial layers).
This localized activation yields a critical insight: the heavy lifting of logical reasoning is predominantly handled by the MLLM. This observation strongly suggests that activating the MLLM during training is essential to fully unleash the model's visual reasoning potential.
This finding directly informs our design choice to \textit{jointly fine-tune} both the MLLM and DiT components in \methodname.

\noindent \textbf{Limited Single-Step Reasoning.}
While the MLLM handles the bulk of logical reasoning, we observe that a single forward pass through the MLLM is insufficient for complex tasks.
As depicted in Fig.\ref{fig:analysis/fail}(b), in lower-complexity scenarios (e.g., 8$\times$8 mazes), the DiT successfully focuses on the generated trajectory. However, the generated paths systematically violate physical constraints (e.g., passing through walls).
This indicates that while the DiT successfully executes spatial grounding effectively in simple tasks, the MLLM lacks the capacity to fully encode and enforce all logical constraints in a single pass.
The MLLM needs to iteratively refine its understanding of the problem, rather than attempting to compute the complete solution in one shot.
This motivates our design of \textit{multi-round reasoning} in the latent space.

\noindent \textbf{Static-Guidance Failure in Complex Dynamic DiT Decoding.} Even if the MLLM could produce perfect reasoning in a single step, we observe a second critical limitation: the information coupling between the DiT and MLLM is fragile and static.
To diagnose this, we analyze the cross-attention entropy between the generated spatial patches and the ground-truth reasoning tokens during the denoising process in Fig.~\ref{fig:analysis/fail}(c).
In high-complexity scenarios (e.g., 32$\times$32 mazes), the attention entropy map becomes globally high and diffuse. This reveals a severe breakdown in the coupling between the DiT and the MLLM. When faced with dense spatial topologies, the DiT loses its ability to anchor spatial features to specific logical text tokens, causing the attention distribution to average out and resulting in a complete collapse of spatial grounding.
The static, one-time injection of text embeddings is insufficient, and the conditioning signal needs to be \textit{dynamically updated} throughout the reasoning process.



\noindent \textbf{Summary.} Our analysis across Fig.~\ref{fig:analysis/fail}(a-c) highlights three key insights: (1) The MLLM has strong reasoning potential, but cannot fully exploit it in a single forward pass; (2) The DiT excels at spatial grounding, but requires dynamic, evolving conditioning to maintain alignment with complex logical constraints; (3) The current paradigm of static, one-time text injection is limited for multi-step reasoning tasks.
This insight motivates our iterative latent reasoning mechanism, which we present in the next section.
\section{Methods}

\begin{figure}[t]
        \centering
        \includegraphics[width=1\linewidth]{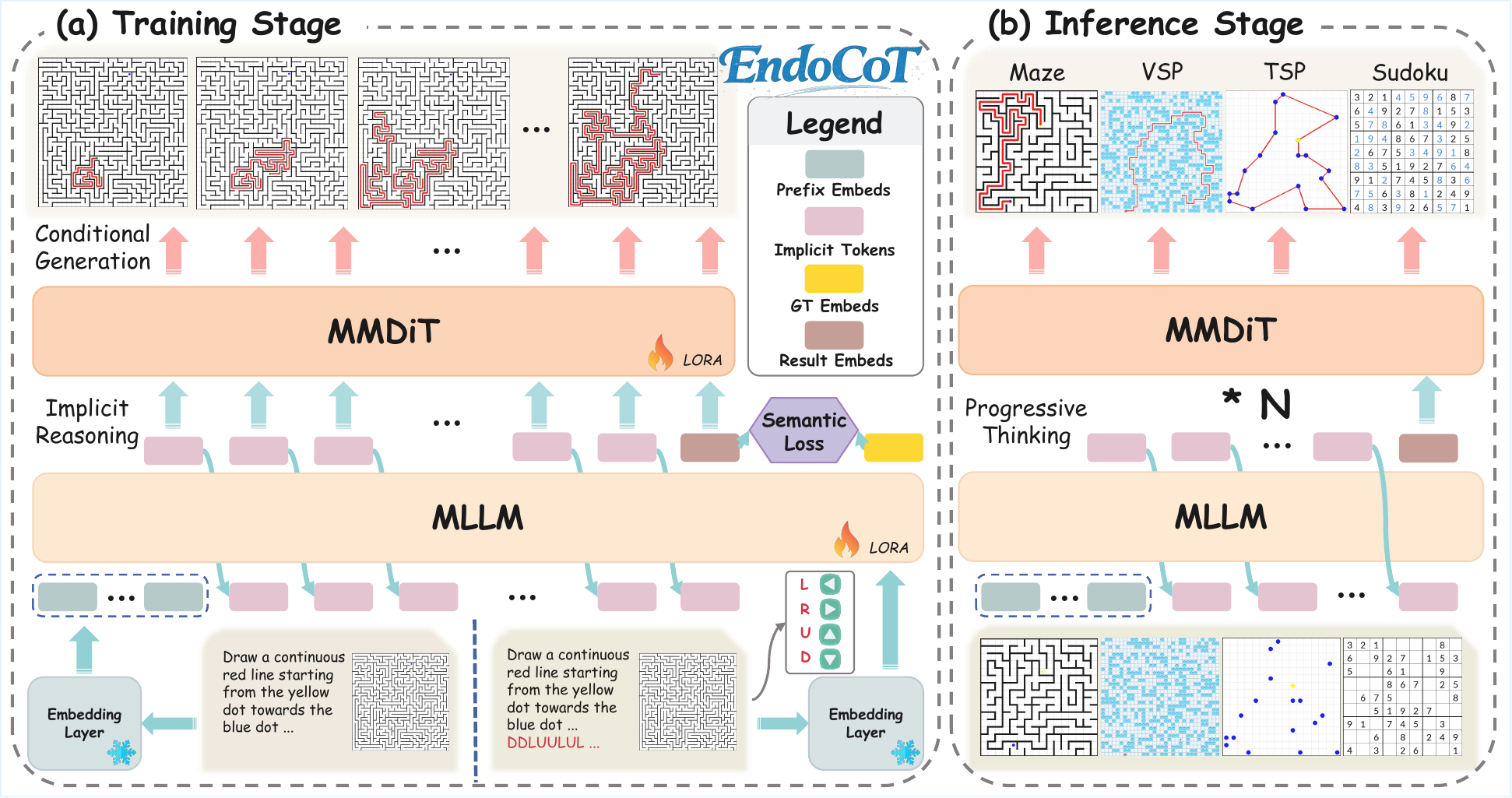}
        \vspace{-12pt}
        \caption{\small \textbf{Overview of \methodname}. (a) Training: We propose a progressive two-stage training strategy: the first stage trains the model to fit both intermediate and final states at each reasoning step, capturing the full multi-step trajectory; the second stage freezes gradients on intermediate states and optimizes only the terminal state, refining generation quality while preserving learned reasoning dynamics (b) Inference:  the model iteratively updates latent representations.}
        \label{fig:main}
        \vspace{-12pt}
\end{figure}

Building on our analysis in Sec. \ref{sec:analysis}, we present \textit{\textbf{Endo}genous} \textbf{C}hain-\textbf{o}f-\textbf{T}hought (\textbf{\methodname}), a framework that enables diffusion models to perform self-guided reasoning during generation.
As shown in Fig.~\ref{fig:main}, our approach iteratively refines latent thought states in the MLLM to create a CoT reasoning process, aligns these reasoning states with explicit textual supervision, and uses progressive training to first build reasoning capability, then refine output quality.

We begin with necessary preliminaries on flow matching, then present our core methodology in three parts: (1) iterative thought guidance module, (2) terminal thought grounding, and (3) progressive training strategy. We conclude with the full inference algorithm.
\subsection{Preliminary}
\noindent \textbf{Flow Matching (FM).}
Flow Matching \cite{lipman2022flow} provides a simplified framework for generative modeling by constructing a linear probability path between two distributions.
Let $X_0 \sim \pi_0$ denote a clean data sample and $X_1 \sim \pi_1$ represent Gaussian noise from the prior.
A linear trajectory $X_t$ interpolates between $X_0$ and $X_1$ over a continuous time step $t \in [0, 1]$:
\begin{equation}
    X_t = t X_1 + (1 - t) X_0.     
\end{equation}
By differentiating $X_t$ with respect to time step $t$, we obtain the ground-truth vector field $u_t(X_t)$: 
\begin{equation}
    u_t(X_t) = \frac{dX_t}{dt} = X_1 - X_0.
\end{equation}
A neural network $v_\theta(X_t, t, c)$ approximates this vector field, where $c$ is an optional conditioning variable.
The training objective $\mathcal{L}_{\text{FM}}$ is :
\begin{equation}
    \label{eq:basic}
    \mathcal{L}_{\text{FM}} = \mathbb{E}_{t, X_0 \sim \pi_0, X_1 \sim \pi_1}  \left\| v_\theta(X_t, t, c) - u_t(X_t) \right\|^2.
\end{equation}

\subsection{\methodname}
Current diffusion models exhibit limited reasoning capabilities for complex, multi-step tasks. Recent analyses suggest these models possess reasoning potential matching their parameter scale. Architectural advances in text encoders \cite{blackforestlabs2025flux2, wu2025qwen} provide the structural foundation to unlock this potential. We propose \textit{Endogenous} Chain-of-Thought (\textbf{\methodname}), a framework enabling diffusion models to perform autonomous chain-of-thought reasoning.


\subsubsection{Iterative Thought Guidance.}
\label{sec:reasoning_conditional}

\begin{figure}[t]
    \centering
    \includegraphics[width=1\linewidth]{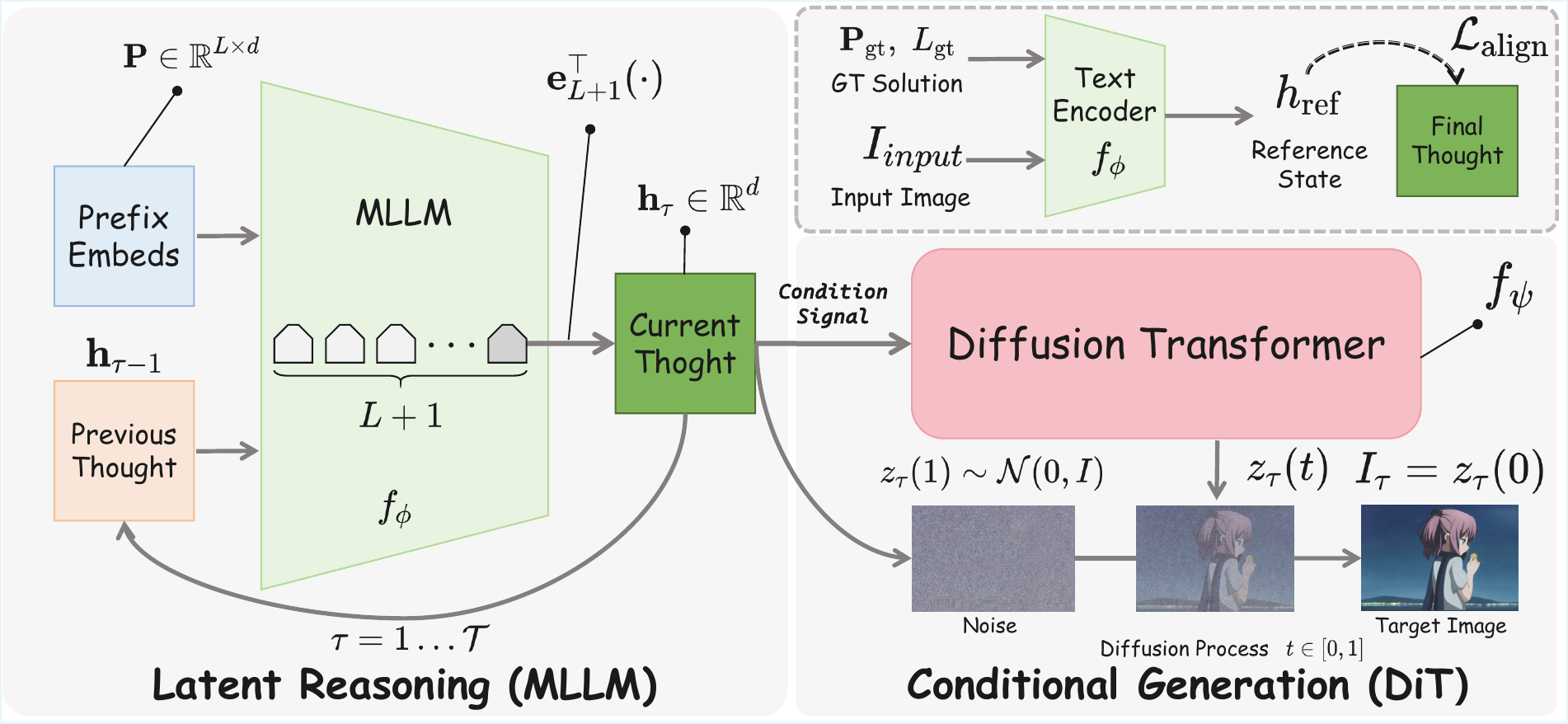}
    \vspace{-12pt}
    \caption{\textbf{Overview of notations and iterative thought guidance module.} \methodname iteratively refines latent states \textcolor{blue}{$\mathbf{h}_\tau$} through the MLLM $f_\phi$, then conditions the DiT $f_\psi$ at each reasoning step $\tau$ to generate intermediate visual outputs $\mathbf{I}_\tau$.}
    \label{fig:notation}
    \vspace{-12pt}
\end{figure}

We formulate the iterative thought guidance process as an iterative refinement of conditional latent states. Standard diffusion models generate an output by conditioning on static text embeddings. In contrast, our approach performs $\mathcal{T}$ reasoning steps, where each step refines both the visual output and the conditioning signal.

Intuitively, this process mimics human problem-solving: rather than attempting to generate the complete solution in one pass, we iteratively refine our understanding of the problem and proposed solution.
Each reasoning step builds upon the previous one, allowing the model to explore solution spaces incrementally.

Let $f_\phi$ denote the MLLM (excluding embedding and projection layers) and $f_\psi$ denote the Diffusion Transformer (DiT). A summary of the main notations is provided in Fig.~\ref{fig:notation}.
The reasoning process is formulated as a sequence of hidden state updates in the latent manifold $\mathbb{R}^d$. $\mathbf{P} \in \mathbb{R}^{L \times d}$ represents the prefix embeddings obtained by encoding the textual prompt and input image through their respective embedding layers. Given fixed prefix embeddings $\mathbf{P}$ (where $L$ is the sequence length of the textual prompt), the $\tau$-th reasoning step ($\tau \in \{1, \dots, \mathcal{T}\}$) updates the thought state $\textcolor{blue}{\mathbf{h}_\tau} \in \mathbb{R}^d$ recursively:
\begin{equation}\label{eq:latent}
    \textcolor{blue}{\mathbf{h}_\tau} = \mathbf{e}_{L+1}^\top f_\phi\left( [\mathbf{P} ; \textcolor{blue}{\mathbf{h}_{\tau-1}}] \right), \quad \tau = 1, \dots, \mathcal{T},
\end{equation}
where $[\cdot ; \cdot]$ denotes concatenation along the sequence dimension, and $\mathbf{e}_{L+1}$ is the one-hot basis vector used to extract the hidden state at the $(L+1)$-th sequence position.
Crucially, $\textcolor{blue}{\mathbf{h}_{\tau-1}}$ directly serves as a high-dimensional input to the first layer of $f_\phi$, bypassing the discrete embedding lookup table.

\noindent \textbf{Conditional Flow Generation.}
Note that our reasoning steps $\mathcal{T}$ are distinct from the diffusion model's internal denoising timesteps.
Each reasoning step $\tau$ involves a complete denoising trajectory from noise to image, conditioned on the current thought state $\textcolor{blue}{\mathbf{h}_\tau}$.
At each reasoning step $\tau$, we condition the diffusion model on the current thought state $\textcolor{blue}{\mathbf{h}_\tau}$ to generate an intermediate visual output $\mathbf{I}_\tau$ by solving the flow ODE:
\begin{equation}
\frac{d\mathbf{z}_\tau(t)}{dt} = v_\psi(\mathbf{z}_\tau(t), t, \textcolor{blue}{\mathbf{h}_\tau}), \quad \mathbf{z}_\tau(1) \sim \mathcal{N}(\mathbf{0}, \mathbf{\textit{I}}), \quad \mathbf{I}_\tau = \mathbf{z}_\tau(0),
\end{equation}
where $\mathbf{z}_\tau(t)$ denotes the latent state at flow time $t \in [0, 1]$. We optimize the model by supervising the generated output against a ground-truth intermediate target $\mathbf{I}_\tau^*$, using the Conditional Flow Matching loss based on Eq. \ref{eq:basic}:
\begin{equation}
    \mathcal{L}_{\text{reasoning}} = \mathbb{E}_{\tau, t, \mathbf{z}_{\tau}(0), \mathbf{z}_{\tau}(1)} \left[ \| (\mathbf{z}_{\tau}(0) - \mathbf{z}_{\tau}(1)) - v_\psi(\mathbf{z}_\tau(t), t, \textcolor{blue}{\mathbf{h}_\tau}) \|^2 \right].
\end{equation}
The intermediate targets $\mathbf{I}_\tau^*$ can be obtained through sequential ground-truth decomposition (e.g., partial path segments for mazes).

Motivated by our analysis in Sec. \ref{sec:analysis}, we apply LoRA fine-tuning to both $f_\phi$ and $f_\psi$. This joint adaptation enables collaborative reasoning across the conditioning and generation stages.
\subsubsection{Terminal Thought Grounding.}\label{sec:text_alignment}
While the reasoning-enhanced framework enables iterative visual refinement, its reliance on purely visual supervision introduces two primary challenges: (1) a modality gap between visual targets and latent reasoning states, and (2) underutilization of explicit textual ground truth.
Inspired by the success of explicit supervision in text-based latent reasoning, we introduce an auxiliary alignment objective that grounds the final reasoning state with textual supervision:
\begin{equation}
    \textcolor{blue}{\mathbf{h}_{\text{ref}}} = \mathbf{e}_{L_{\text{gt}}+1}^\top f_\phi\left( [\mathbf{P}_{\text{gt}} , \mathbf{I}_{\text{input}}] \right),
    \end{equation}
where $\mathbf{P}_{\text{gt}} \in \mathbb{R}^{L_{\text{gt}} \times d}$ is the embedding of the ground-truth reasoning steps, and $\mathbf{I}_{\text{input}}$ denotes the task input (e.g., initial maze configuration).
We align the final reasoning state $\textcolor{blue}{\mathbf{h}_\mathcal{T}}$ with this reference using an L2 (Semantic) loss:
\begin{equation}
\mathcal{L}_{\text{align}} = \| \textcolor{blue}{\mathbf{h}_\mathcal{T}} - \textcolor{blue}{\mathbf{h}_{\text{ref}}} \|^2.
\end{equation}
The overall training loss combines the flow matching loss with this alignment term:
\begin{equation}\label{eq:loss}
    \mathcal{L}_{\text{total}} = \mathcal{L}_{\text{FM}} + \mathbb{I}_{\{\tau = \mathcal{T}\}} \cdot \lambda_{\text{align}} \mathcal{L}_{\text{align}},
\end{equation}
where $\mathbb{I}_{\{\tau = \mathcal{T}\}}$ is the indicator function activating the alignment loss only at the final reasoning step, and $\lambda_{\text{align}}$ balances visual generation quality and textual grounding.
Empirically, we set $\lambda_{\text{align}} = 1$.
As shown in our ablation study (Sec.~\ref{sec:ablation}), this term is critical for preventing drift in the latent reasoning states and ensuring the reasoning trajectory remains grounded.

\subsubsection{Progressive Training Strategy} \label{sec:progressive_training}
Reasoning steps and final outputs serve distinct objectives: intermediate steps focus on exploring solution paths, while the final step emphasizes producing a correct answer. Directly optimizing both objectives simultaneously can lead to conflicting gradients. We therefore adopt a two-stage progressive training strategy.

\noindent \textbf{Stage 1: Reasoning Development.}
As shown in Fig. \ref{fig:main} (a), we supervise all reasoning steps $\tau = 1, \ldots, \mathcal{T}$, enabling the model to learn step-by-step visual reasoning.
The training objective is: 
\begin{equation}
\mathcal{L}_{\text{stage1}} = \sum_{\tau=1}^{\mathcal{T}} \left( \mathcal{L}_{\text{FM}}^{\tau} + \mathbb{I}_{\{\tau = \mathcal{T}\}} \lambda_{\text{align}} \mathcal{L}_{\text{align}} \right),
\end{equation}
where $\mathcal{L}_{\text{FM}}^{\tau}$ denotes the flow matching loss at reasoning step $\tau$.
By providing supervision at each intermediate step, we encourage the model to develop coherent, incremental reasoning trajectories.

\noindent \textbf{Stage 2: Terminal Consolidation.}
Once the model has developed robust reasoning capabilities, we shift the training focus to solidifying the visual accuracy of the final output.
As shown in Fig. \ref{fig:main} (b), while the intermediate reasoning process is preserved during the forward pass, gradients are computed exclusively for the final output:

\vspace{1em}
\begin{equation}
\mathcal{L}_{\text{stage2}} = \mathcal{L}_{\text{FM}}^{\mathcal{T}} + \lambda_{\text{align}} \mathcal{L}_{\text{align}}.
\end{equation}

The intermediate steps $\tau < \mathcal{T}$ are executed in the forward pass only, serving as reasoning scaffolding without gradient propagation. To prevent the degradation of the previously learned reasoning chains, Stage 2 employs a short-cycle fine-tuning strategy with limited iterations.

\vspace{-1.5em}

\subsubsection{Inference Process} 
As shown in Fig. \ref{fig:main}(b), \methodname does not require decoding intermediate visual states during inference. By specifying the number of reasoning steps $\mathcal{T}$, the model recursively updates its latent thought states to generate the final result.

\vspace{-1em}

\section{Experiments}\label{sec:experiments}

\begin{figure}[t]
    \centering
    \includegraphics[width=1\linewidth]{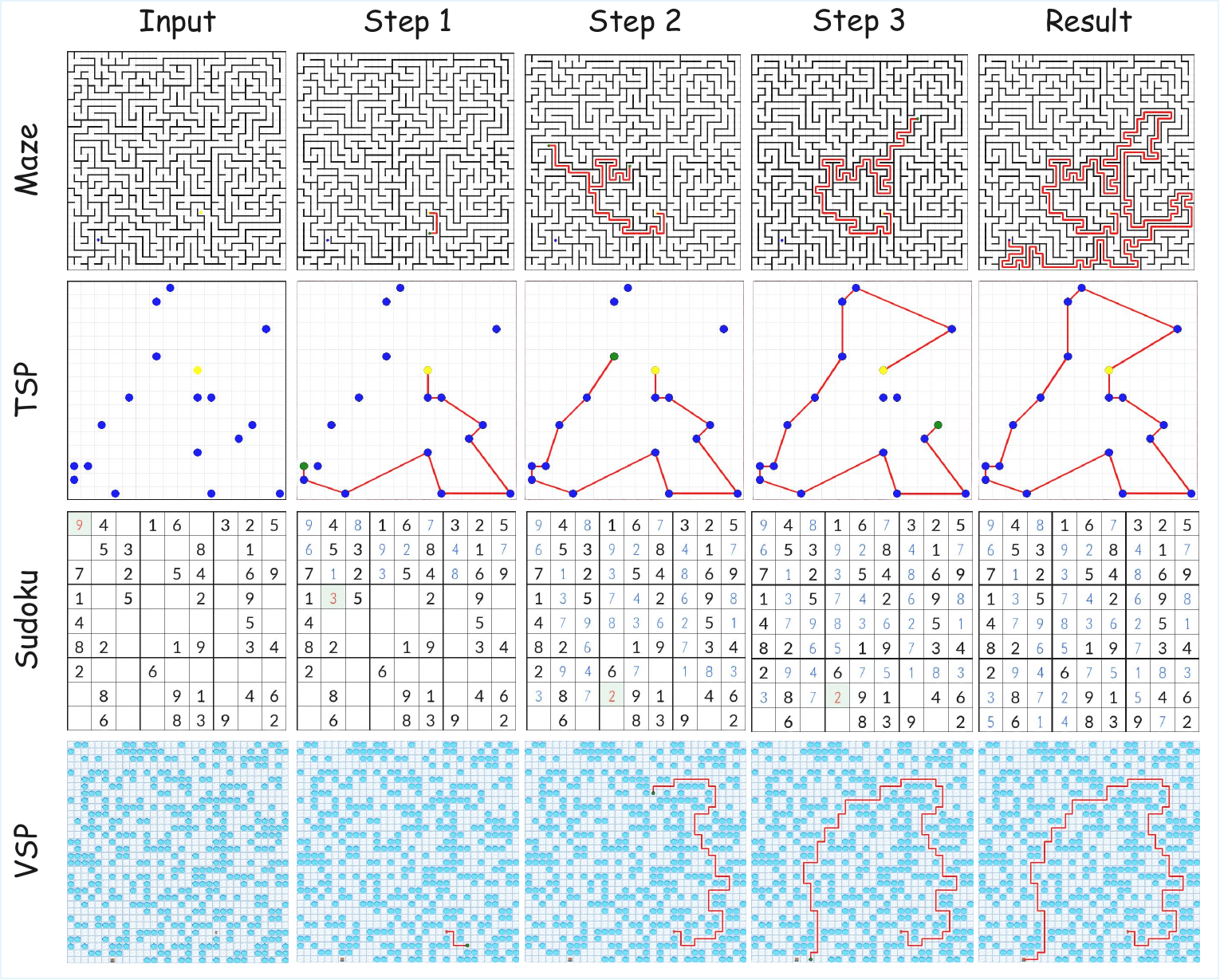}
    \vspace{-12pt}
    \caption{\small
    \textbf{\footnotesize Step-by-step reasoning process across four distinct tasks.}
    Our model incrementally resolves complex visual reasoning tasks through intermediate reasoning steps.
    For each task (Maze, TSP, Sudoku, VSP), we show the initial input (leftmost), intermediate refinement steps, and the final optimal solution (rightmost).}
    \label{fig:examples}
    \vspace{-20pt}
    \label{fig:experiments/case}
\end{figure}

\noindent \textbf{Model Setup.}
We build upon Qwen-Image-Edit-2511 \cite{wu2025qwen} and apply LoRA fine-tuning \cite{hu2022lora} to efficiently adapt the model while preserving pre-trained knowledge.
We use a LoRA rank of 32 for all targeted modules, a learning rate of $1 \times 10^{-4}$, and train for $5$ epochs.
Detailed hyperparameter and LoRA module configuration are provided in the Appendix.
    
\subsection{Results on Visual Reasoning Tasks}
\noindent \textbf{Datasets.}
Following the evaluation protocol of \cite{he2025diffthinker}, we select Maze navigation (Maze), Traveling Salesman Problem (TSP), Sudoku solving (Sudoku), Visual Spatial Planning (VSP), and VSP-Pro with larger map sizes (up to $32 \times 32$) as evaluation benchmarks.
Each task requires multi-step logical reasoning.
We also follow \cite{he2025diffthinker} to construct the fine-tuning data of each task. All generated data are rendered at a resolution of $512 \times 512$.
Detailed statistics and construction methods of fine-tuning data are provided in the Appendix.

\noindent \textbf{Settings.} We evaluate models under three distinct settings.
\textit{1) Zero-shot:} Vanilla baselines without any task-specific training.
\textit{2) Task-specific training:} Training and evaluation are performed on individual tasks separately (e.g., train on the Maze dataset, evaluate on Maze only), which is the default setting of DiffThinker \cite{he2025diffthinker}.
\textit{3) Unified training:} A single model is trained on the combined dataset of all tasks (Maze+TSP+Sudoku+VSP), and evaluated on all tasks simultaneously.
The `unified training' setting is more challenging as it requires the model to handle a wide range of problems.

\begin{table}[t]
    \centering
    \caption{\small \textbf{Evaluation results across visual reasoning tasks}. Our model establishes a new state-of-the-art across both task-specific and unified training settings, demonstrating exceptional robustness to increasing task complexity.}
    \vspace{-12pt}
    \label{tab:experiment/main}
    \resizebox{1.\linewidth}{!}{
    \begin{tabular}{l ccc ccc cccc cccccccc c}
    \toprule
    \multirow{2}{*}{\textbf{Models}} & \multicolumn{3}{c}{\textbf{Maze}} & \multicolumn{3}{c}{\textbf{TSP}} & \multicolumn{4}{c}{\textbf{Sudoku}} & \multicolumn{6}{c}{\textbf{VSP}} & \multicolumn{2}{c}{\textbf{VSP-Super}} & \multirow{2}{*}{\textbf{Avg}} \\
    \cmidrule(r){2-4} \cmidrule(r){5-7} \cmidrule(r){8-11} \cmidrule(r){12-17} \cmidrule(r){18-19}
    ~ & 8 & 16 & 32 & 12 & 15 & 18 & 45 & 40 & 35 & 30 & 3 & 4 & 5 & 6 & 7 & 8 & 16 & 32 \\
    \cmidrule(r){1-1} \cmidrule(r){2-4} \cmidrule(r){5-7} \cmidrule(r){8-11} \cmidrule(r){12-17} \cmidrule(r){18-19} \cmidrule(r){20-20}
    \rowcolor{gray!18} \multicolumn{20}{l}{$\blacktriangledown$ \emph{Zero-Shot}} \\
    ThinkGen \cite{jiao2025thinkgen} & 0  & 0 & 0 & 0 & 0 & 0 & 0 & 0 & 0 & 0 & 44 & 4 & 11 & 13 & 9 & 10 & 0 & 0 & 5.1\\
    ChronoEdit \cite{wu2025chronoedit} & 1  & 1 & 26 & 0 & 0 & 0 & 0 & 0 & 0 & 0 & 60 & 20 & 12 & 7 & 16 & 13 & 1 & 1 & 8.8\\
    Qwen3-VL-8B \cite{yang2025qwen3} & 1 & 0 & 0  & 0 & 0 & 0 & 0 & 0 & 0 & 0& 64 & 46 & 33 & 21 & 12 & 21 & 1 & 0 & 11.1\\
    Qwen-Image-Edit-2511 \cite{wu2025qwen} & 0  & 0 & 0  & 0 & 0 & 0 & 0 & 0 & 0 & 0 & 50 & 55 & 44 & 16 & 23 & 23 & 0 & 0 & 11.7\\
    \cmidrule(r){1-1} \cmidrule(r){2-4} \cmidrule(r){5-7} \cmidrule(r){8-11} \cmidrule(r){12-17} \cmidrule(r){18-19} \cmidrule(r){20-20}
    \rowcolor{gray!18} \multicolumn{20}{l}{$\blacktriangledown$ \emph{Task-Specific Training}} \\
    Qwen3-VL-8B(SFT) \cite{he2025diffthinker} & 53  & 37 & 0  & 59 & 60 & 43 & 30 & 17 & 2 & 0 & 99 & 96 & 98 & 96 & 92 & 86 & 61 & 8 & 58.56\\
    Qwen3-VL-8B(GRPO) \cite{he2025diffthinker} & 0  & 0 & 0  & 0 & 0 & 0 & 0 & 0 & 0 & 0 & 91 & 70 & 70 & 31 & 34 & 24 & 0 & 0 & 17.8 \\
    DiffThinker \cite{he2025diffthinker} & \textbf{100} & \textbf{100} & 65 & 76 & 72 & 59 & 97 & 94 & 55 & 13 & \textbf{100} & \textbf{100} & \textbf{100} & 98 & \textbf{100} & \textbf{100} & 99 & 80 & 83.8\\
    \rowcolor[HTML]{E2F4E3}
    \textbf{Ours} & \textbf{100} & \textbf{100} & \textbf{90} & \textbf{77} & \textbf{77} & \textbf{73} & \textbf{100} & \textbf{100} & \textbf{95} & \textbf{64} & \textbf{100} & \textbf{100} & \textbf{100} & 99 & \textbf{100} & 99 & 99 & \textbf{85} & \textbf{92.1}\\
    \cmidrule(r){1-1} \cmidrule(r){2-4} \cmidrule(r){5-7} \cmidrule(r){8-11} \cmidrule(r){12-17} \cmidrule(r){18-19} \cmidrule(r){20-20}
    \rowcolor{gray!18} \multicolumn{20}{l}{$\blacktriangledown$ \emph{Unified Training}} \\
    DiffThinker \cite{he2025diffthinker} & \textbf{98} & \textbf{99} & \textbf{66} & \textbf{64} & 49 & 34 & 94 & 45 & 26 & 37 & \textbf{100} & \textbf{99} & \textbf{99} & 99 & \textbf{100} & 97 & 97 & \textbf{84} & 77.1 \\
    \rowcolor[HTML]{E2F4E3} \textbf{Ours} & 97 & 98 & 52 & \textbf{64} & \textbf{55} & \textbf{46} & \textbf{100} & \textbf{88} & \textbf{80} & \textbf{59} & \textbf{100} & 98 & 98 & \textbf{100} & \textbf{100} & \textbf{100} & \textbf{100} & 80 & \textbf{84.2} \\
    \bottomrule
    \end{tabular}}
\vspace{-10pt}
\end{table}

\noindent \textbf{Results of Task-Specific Training.} As shown in Tab. \ref{tab:experiment/main}, our model establishes a new state-of-the-art across all evaluated visual reasoning benchmarks under task-specific training.
Our approach demonstrates exceptional robustness to increasing complexity.
At higher difficulty levels like Sudoku (scale 35) and Maze (scale 32), it achieves 95\% and 90\% accuracy respectively, significantly outperforming DiffThinker \cite{he2025diffthinker} (55\% and 65\%).
On spatial planning tasks, our model maintains near-perfect scores across standard VSP levels and reaches 85\% on the most challenging VSP-Super (scale 32), whereas generative baselines like ThinkGen \cite{jiao2025thinkgen} and ChronoEdit \cite{wu2025chronoedit} fail completely.
Fig. \ref{fig:experiments/case} shows the unique intermediate reasoning process of our model, which progressively refines solutions through multiple steps that are absent in prior approaches.

\noindent \textbf{Results of Unified Training.} The unified training results (single model trained on all tasks) show that our framework can learn transferable reasoning skills across tasks, though performance is slightly lower than task-specific training. As shown in Tab. \ref{tab:experiment/main}, under unified training, our approach still achieves competitive performance compared to DiffThinker \cite{he2025diffthinker}.

\subsection{Ablation Study and Analysis}\label{sec:ablation}
To validate our design choices, we conduct comprehensive ablation studies on the Maze benchmark, which serves as a representative testbed for long-horizon reasoning.
We report both the accuracy and \textbf{path repetition rate} that measures the overlap ratio between the generated path and the ground truth path.

\begin{table}[t]
\begin{minipage}{0.51\textwidth}
\caption{
\small 
\textbf{\small Ablation Study on the Maze Benchmark.} 
Removing the auxiliary semantic loss or using explicit tokens both lead to significant performance degradation compared to our default choice.
}
\label{tab:ablation-full}
\vspace{-12pt}
\centering
\resizebox{1.\linewidth}{!}{
\begin{tabular}{l lll lll}
\toprule
\multirow{2.5}{*}{\textbf{Models}} & \multicolumn{3}{c}{\textbf{Accuracy (ACC)}} & \multicolumn{3}{c}{\textbf{Path Repetition (\%)}} \\
\cmidrule(l){2-4} \cmidrule(l){5-7}
& \textbf{8} & \textbf{16} & \textbf{32} & \textbf{8} & \textbf{16} & \textbf{32} \\
\cmidrule(l){1-1} \cmidrule(l){2-4} \cmidrule(l){5-7}
w/o semantic loss & 39 & 42 & 14 & 93.44 & 92.23 & 67.24 \\
Explicit token  & 34 & 8  & 0  & 81.47 & 33.35 & 0.08  \\
\midrule
\rowcolor[HTML]{E2F4E3} 
\textbf{Ours} & \textbf{100} & \textbf{100} & \textbf{90} & \textbf{100.00} & \textbf{100.00} & \textbf{98.13} \\
\bottomrule
\end{tabular}}
\end{minipage}
\vspace{-6pt}
\hspace{+1pt}
\begin{minipage}{0.48\textwidth}
\caption{
\small
\textbf{\small Performance comparison across different maze scales.}
Our method achieves near-perfect accuracy and correct path repetition rate, significantly surpassing the baselines.
}
\vspace{-12pt}
\resizebox{1.\linewidth}{!}{
\begin{tabular}{l lll lll}
\toprule
\multirow{2}{*}{\textbf{Models}} & \multicolumn{3}{c}{\textbf{Accuracy}} & \multicolumn{3}{c}{\textbf{Path Repetition (\%)}} \\
\cmidrule(r){2-4} \cmidrule(r){5-7}
~ & \textbf{8} & \textbf{16} & \textbf{32} & \textbf{8} & \textbf{16} & \textbf{32} \\
\cmidrule(r){1-1} \cmidrule(r){2-4} \cmidrule(r){5-7}
MLLM-Only & 0  & 0  & 0  & 0.00  & 0.62  & 0.20  \\
DiT-Only  & 57 & 43 & 18 & 89.75 & 90.80 & 80.96 \\
\rowcolor[HTML]{E2F4E3}
\textbf{Ours} & \textbf{100} & \textbf{100} & \textbf{90} & \textbf{100.00} & \textbf{100.00} & \textbf{98.13} \\
\bottomrule
\end{tabular}}
\label{tab:maze-main-results}
\end{minipage}
\vspace{-6pt}
\end{table}

\noindent \textbf{Effectiveness of the Semantic Loss.}
The auxiliary Semantic loss provides explicit textual supervision by aligning our continuous latent token representations with ground-truth reasoning steps.
This constraint guides the implicit tokens toward semantically meaningful trajectories. 

As shown in Tab. \ref{tab:ablation-full}, removing this supervision signal (\textit{w/o} semantic loss) leads to severe performance degradation.
While the model retains basic routing ability on simple mazes, its logical consistency breaks down on harder instances: on the highly complex Maze-32 benchmark, accuracy drops from 90\% to just 14\%, and path repetition rate falls from 98.13\% to 67.24\%.
This confirms that the semantic loss acts as a crucial regularizer, preventing implicit token drift and enabling reliable long-horizon reasoning.

\noindent \textbf{Implicit vs. Explicit Tokens.}
While continuous implicit tokens offer flexibility, one might reasonably consider explicit text generation as a more interpretable alternative for reasoning \cite{shen2025codi,hao2024training}.
To investigate this design choice, we construct a variant that replaces implicit tokens with explicit autoregressive text generation, where the MLLM must produce discrete intermediate reasoning steps before generating the final visual output.

\begin{table}[t]
\caption{
\small
\textbf{Inference-Time CoT Scaling on Maze Benchmarks.}
Comparison of Accuracy and Path Repetition Rate across different token budgets ($\tau$). Performance consistently improves, particularly on challenging Maze-32.
}
\label{tab:maze-scaling-combined}
\vspace{-12pt}
\centering
\small
\resizebox{.7\linewidth}{!}{
\begin{tabularx}{\columnwidth}{l *{7}{>{\centering\arraybackslash}X}}
                \toprule
                \multirow{2}{*}{\textbf{Models}} & \multicolumn{3}{c}{\textbf{Accuracy}} & \multicolumn{3}{c}{\textbf{Path Repetition (\%)}} & \multirow{2}{*}{\textbf{Time}} \\
                \cmidrule(lr){2-4} \cmidrule(lr){5-7}
                & \textbf{8} & \textbf{16} & \textbf{32} & \textbf{8} & \textbf{16} & \textbf{32} & \\
                \midrule
                DiffThinker \cite{he2025diffthinker} & 100 & 97 & 56 & 100.00 & 100.00 & 73.74 & 15.72 \\
                \midrule
                Ours ($\tau=2$)  & 100 & 72  & 11 & 100.00 & 89.15 & 45.26 & 16.02 \\
                Ours ($\tau=5$) & 100 & 94  & 27 & 100.00 & 98.49 & 63.90 & 16.54 \\
                Ours ($\tau=10$) & 100 & 99  & 49 & 100.00 & 99.93 & 82.33 & 17.52 \\
                Ours ($\tau=20$) & 100 & 100 & 74 & 100.00 & 100.00 & 96.47 & 19.27 \\
                \rowcolor[HTML]{E2F4E3}
                \textbf{Ours ($\tau=50$)} & \textbf{100} & \textbf{100} & \textbf{90} & \textbf{100.00} & \textbf{100.00} & \textbf{98.13} & 24.81 \\
                \bottomrule
\end{tabularx}}
\vspace{-12pt}
\end{table}

\begin{figure}[t]
    \begin{minipage}[t]{0.42\textwidth} 
    \includegraphics[width=\linewidth]{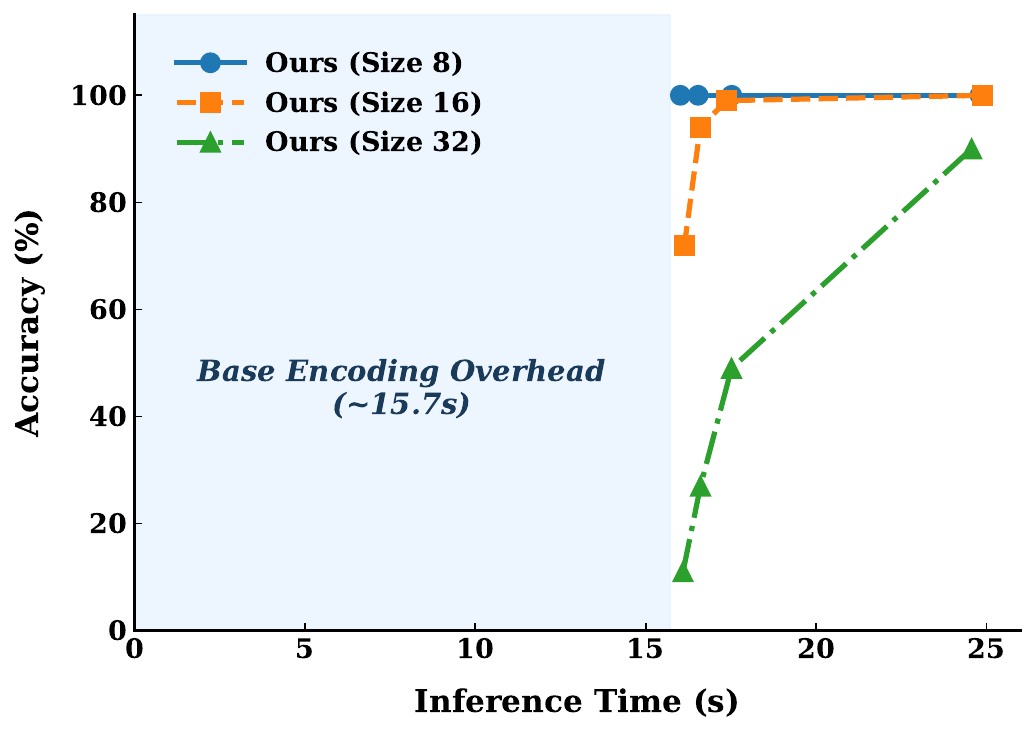}
    \vspace{-18pt}
    \caption{\small The trade-off between model accuracy and overall inference time across different token budgets.}
    \label{fig:efficacy_reasoning}
    \vspace{-18pt}
    \end{minipage}
    \hspace{0.01\textwidth} 
    \begin{minipage}[t]{0.48\textwidth} 
    \includegraphics[width=\linewidth]{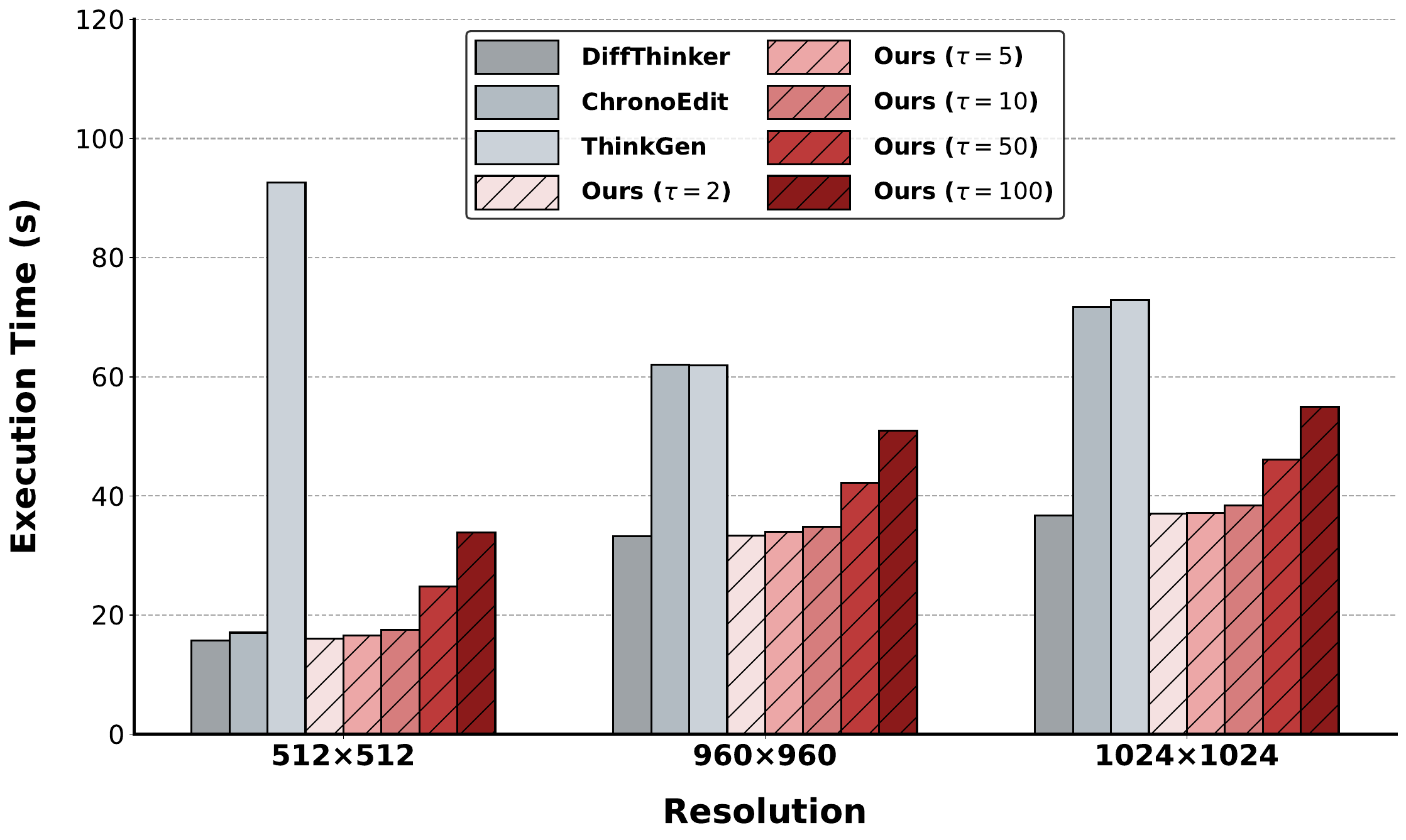}
    \vspace{-18pt}
    \caption{\small Comparison of execution time against existing baselines across different resolutions and token budgets. 
    }
    \label{fig:reasoning_analysis}
    \vspace{-28pt}
    \end{minipage}
\end{figure}

As reported in Tab. \ref{tab:ablation-full}, explicit tokens lead to severe performance degradation.
While the model achieves 34\% accuracy on Maze-8, it fails on Maze-32 (0\% accuracy).
The failure case above reveals the cause: when planning long-horizon paths with a discrete vocabulary, the model becomes vulnerable to autoregressive error accumulation and mode collapse, degenerating into repetitive token loops.

\begin{tcolorbox}[colback=white, colframe=gray!50!black, title={Failure Case of Explicit Tokens}, breakable]
"$\langle \dots \rangle$ <|im\_start|>system\textbackslash n%
                Describe the key features of the input image $\langle \dots \rangle$ Generate a new image that meets the user's requirements $\langle \dots \rangle$ Direction keys: D=Down, U=Up, R=Right, L=Left.<|im\_end|>\textbackslash n%
                <|im\_start|>assistant\textbackslash n%
                \textbf{\textit{\textcolor{failred}{Rencontre Rencontre Rencontre $\langle \dots \rangle$ Rencontre Rencontre}"}}
\end{tcolorbox}

\noindent \textbf{Inference-Time CoT Scaling.} As shown in Tab. \ref{tab:maze-scaling-combined}, our approach demonstrates a robust scaling law: dynamically increasing the number of implicit reasoning budgets $\tau$ smoothly and significantly elevates both accuracy and path repetition rate, particularly on the most challenging \textit{Maze-32} benchmark.
This indicates that our implicit tokens effectively refine complex reasoning steps through iterative exploration.
Furthermore, true scalability requires predictable computational costs. As shown in Fig. \ref{fig:efficacy_reasoning}, our model steadily trades inference time for higher accuracy.

\noindent \textbf{Resolution Scaling.} Beyond the token budget $\tau$, our model demonstrates a crucial advantage in high-resolution tasks: \textit{as spatial resolution increases, the relative computational cost of our method significantly decreases}.
This efficiency gain stems fundamentally from the fact that our approach does not require repeating the computationally expensive DiT denoising steps.
As shown in Fig. \ref{fig:reasoning_analysis}, our model maintains a stable and predictable inference latency.
            
\noindent \textbf{Joint Training vs. MLLM-Only Baselines.}

Finally, we investigate the necessity of jointly optimizing both the MLLM and DiT components, rather than relying on either module in isolation.
Our core premise is that spatial reasoning requires both high-level cognitive planning (from the MLLM) and low-level physical grounding (from the DiT).

\begin{figure*}[t]
\centering
\setlength{\tabcolsep}{1pt}
    \begin{tabular}{cc @{\hspace{8pt}} cc @{\hspace{8pt}} cc}
        \multicolumn{2}{c}{\small \textbf{Case 1}} &
        \multicolumn{2}{c}{\small \textbf{Case 2}} &
        \multicolumn{2}{c}{\small \textbf{Case 3}} \\
        \addlinespace[2pt]
    
        \scriptsize Ori & \scriptsize MLLM-Only &
        \scriptsize Ori & \scriptsize MLLM-Only &
        \scriptsize Ori & \scriptsize MLLM-Only \\
        \addlinespace[4pt]
    
        \includegraphics[width=0.145\textwidth]{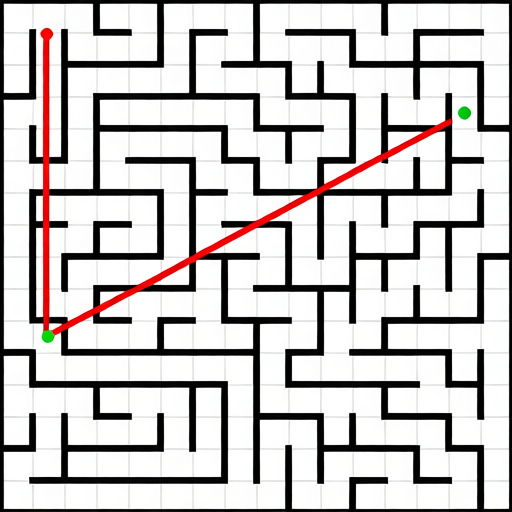} &
        \includegraphics[width=0.145\textwidth]{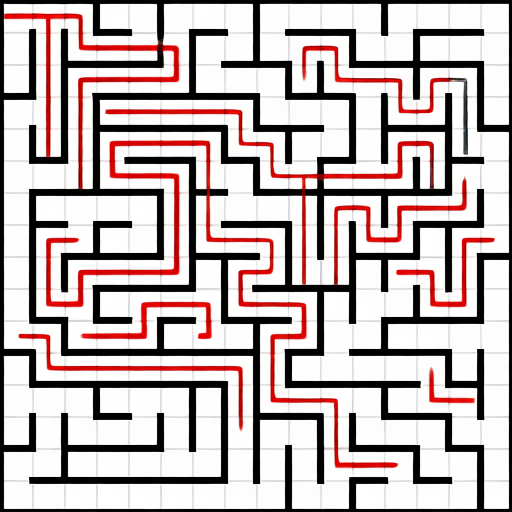} &
        \includegraphics[width=0.145\textwidth]{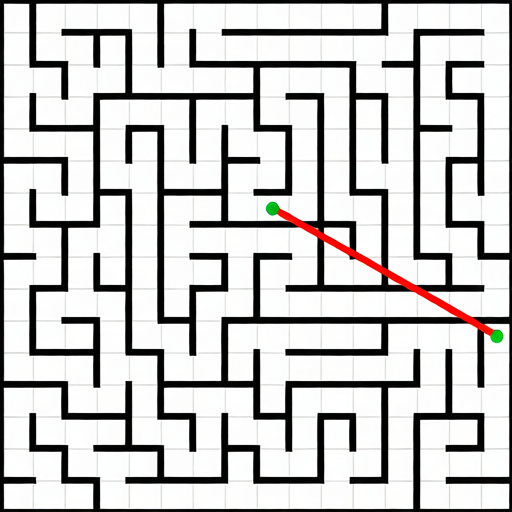} &
        \includegraphics[width=0.145\textwidth]{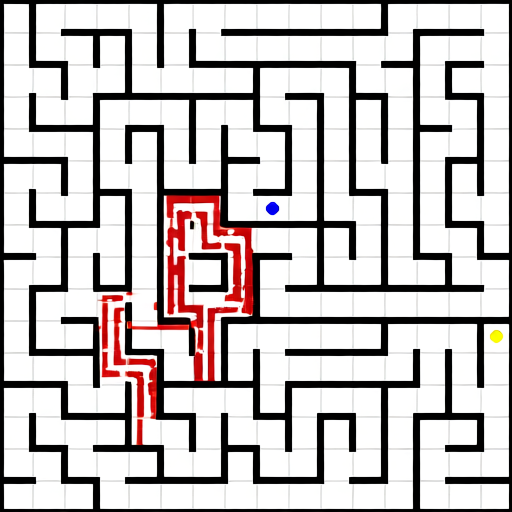} &
        \includegraphics[width=0.145\textwidth]{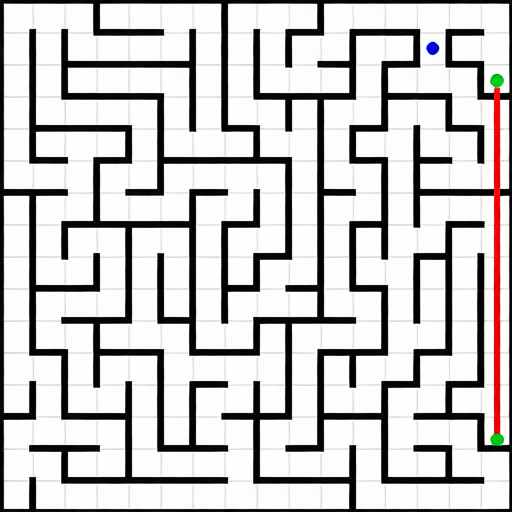} &
        \includegraphics[width=0.145\textwidth]{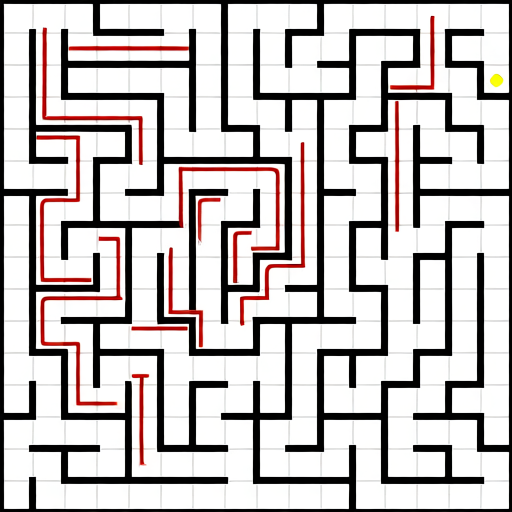} \\
    \end{tabular}
\vspace{-12pt}
\caption{\small
\textbf{Qualitative comparison of path reasoning.} While MLLM-Only enables the model to understand the basic topological structure of the maze compared to the untuned baseline, it still lacks sufficient visual reasoning representations. Consequently, it exhibits erratic and wandering paths.
}
\label{fig:qualitative_maze_wide}
\vspace{-12pt}
\end{figure*}

Quantitatively, Tab. \ref{tab:maze-main-results} shows that decoupling components leads to severe performance degradation.
Using only DiT limits the model's logical reasoning capacity, causing accuracy to drop to 18\% on Maze-32.
Interestingly, while we attribute the model's reasoning capability primarily to the MLLM (Section~\ref{sec:analysis}), relying solely on this module results in complete failure.
This paradox indicates that while language models excel at abstract logic, they cannot directly map conceptual steps into spatial coordinates without the visual grounding provided by the DiT.
This limitation is shown qualitatively in Fig. \ref{fig:qualitative_maze_wide}.
The MLLM-only baseline produces erratic, wandering trajectories that frequently become trapped in dead ends, lacking global spatial awareness.

\noindent \textbf{Results on Image Editing Tasks}

\begin{figure}[t]
    \centering
    \includegraphics[width=1\linewidth]{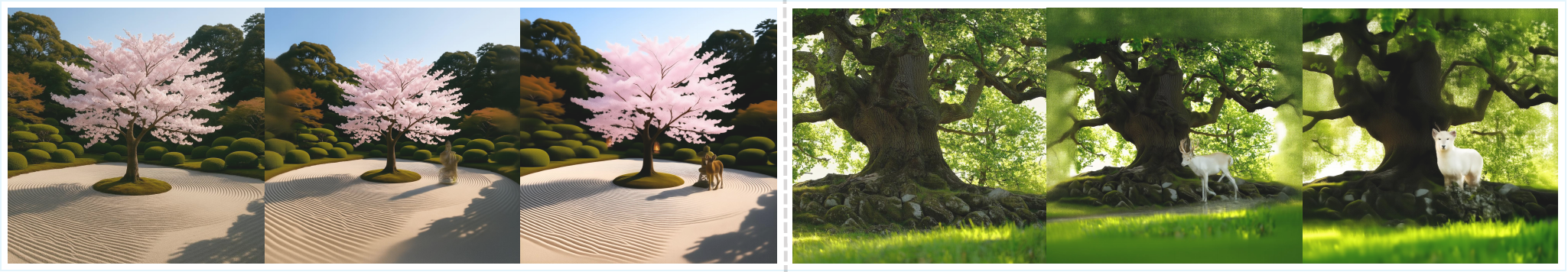}
    \vspace{-18pt}
    \caption{\textbf{Progressive image editing results.} \textbf{Left:} step-by-step object addition, where a stone lantern and a deer are sequentially introduced into the scene. 
    \textbf{Right:} object transformation, where the deer is progressively modified into a sheep.}
    \label{fig:progressive_editing}
    \vspace{-23pt}
\end{figure}

As shown in Fig.~\ref{fig:progressive_editing}, our approach demonstrates a distinctive \textit{progressive editing} capability driven by the internal reasoning trajectory.
Given step-by-step editing instructions, the model iteratively plans and executes each modification during generation.
By varying the reasoning step $\tau$, we can control how many editing operations are carried out.

\section{Conclusion}\label{sec:conclusion}
We presented \textbf{\methodname}, a novel framework that enables diffusion models to perform \textit{endogenous} Chain-of-Thought reasoning. By iteratively refining latent thought states and grounding the final output in textual supervision, EndoCoT successfully bridges the gap between high-level logical planning and precise visual generation. Extensive experiments across diverse reasoning tasks demonstrate that this capability arises from the synergistic coupling of the multimodal text encoder and the DiT backbone, rather than from standard denoising processes alone. While highly effective, our approach currently requires manual tuning for the optimal number of reasoning steps and relies on high-quality datasets with explicit intermediate supervision. Future work will focus on adaptive mechanisms for automatic reasoning depth control and extending the framework to broader general-purpose tasks.

\clearpage
\bibliographystyle{plain}
\bibliography{refs}


\clearpage
\appendix
\newpage
\appendix
\section*{\centering Appendix}

\section{Overview}
In the supplementary material, Sec. \ref{sec:data_statistics} details the dataset statistics; Sec. \ref{sec:case_study} provides task prompts and additional results; Sec. \ref{app:more_ablations} presents additional ablation studies; Sec. \ref{app:training_hints} offers EndoCoT insights and observations; and Sec. \ref{app:edit_data} elaborates on the additional editing datasets.

\section{Data Statistics}
\label{sec:data_statistics}

\begin{table}[h]
    \caption{
        \textbf{Dataset Composition.} Overview of the sample size for each task in our constructed dataset, totaling 182.4K instances.
    }
    \label{tab:dataset-size-horizontal}
    \centering
    \small
    \setlength{\tabcolsep}{8pt}
    \begin{tabular}{l cccc c} 
        \toprule
        \textbf{Task} & Maze & Sudoku & TSP & VSP & \textbf{Total} \\
        \midrule
        \textbf{Size} & 75K & 40K & 30K & 37.4K & \textbf{182.4K} \\
        \bottomrule
    \end{tabular}
\end{table}

    To systematically evaluate our model's reasoning capabilities, we construct a comprehensive dataset totaling 182.4K instances, comprising intermediate reasoning steps. As detailed in Tab. \ref{tab:dataset-size-horizontal}, this dataset spans four distinct reasoning tasks (Maze, TSP, Sudoku, and VSP).

    \subsection{Algorithmic Formulation}
    
    \noindent\textbf{Maze.} We use Depth-First Search (DFS) on a grid to carve a perfectly connected, loop-free maze. After randomly assigning a start ($s$) and goal ($g$), we compute the shortest path $P$ via Breadth-First Search (BFS). This path is recorded iteratively to simulate step-by-step spatial exploration.
    
    \noindent\textbf{TSP.} We randomly distribute $N$ cities on a 2D grid and solve the resulting combinatorial optimization problem using the Held-Karp algorithm. The dynamic programming state transition is defined as:
    \begin{equation} \label{eq:tsp_dp}
        DP[m][i] = \min_{j \in m, j \neq i} \{ DP[m \setminus \{i\}][j] + \text{dist}(j, i) \},
    \end{equation}
    where $m$ is the bitmask of visited cities and $\text{dist}(j, i)$ is the Euclidean distance. The optimal node-to-node trajectory is rendered sequentially.
    
    \noindent\textbf{Sudoku.} We construct a valid $9 \times 9$ board $L$ via backtracking, satisfying standard row, column, and block constraints:
    \begin{equation} \label{eq:sudoku}
        \forall i, \sum_{j=1}^9 L_{i,j} = N; \quad \forall j, \sum_{i=1}^9 L_{i,j} = N; \quad \forall \text{block}_k, \sum L \in \text{block}_k = N.
    \end{equation}
    We iteratively remove clues while ensuring the puzzle retains a unique solution. The intermediate reasoning steps are recorded by reversing this process, progressively filling empty cells from the ground-truth solution.
    
    \noindent\textbf{VSP.} We generate a FrozenLake grid containing safe tiles, hazard holes, a start, and a goal. Framing this as a graph search, we apply Dijkstra's algorithm to compute a safe path. The distance relaxation for adjacent safe nodes $u$ and $v$ is $d(v) = \min \{d(v), d(u) + w(u,v)\}$.
    The agent's state transitions are recorded step-by-step to capture environment perception and hazard avoidance.

    \subsection{Pseudocode for Dataset Generation}

    The procedural generation for all four tasks are shown in Fig. \ref{fig:pseudocode}: \textbf{Maze Generation:} Utilizes \textit{DFS} to construct the maze, followed by \textit{BFS} to determine the optimal path. 
    \textbf{TSP Generation:} Applies the\textit{ Held-Karp dynamic programming algorithm} to derive the optimal tour.
    \textbf{VSP Generation:} Employs \textit{Dijkstra's algorithm} on the remaining safe graph to compute shortest paths.
    \textbf{Sudoku Generation:} Starts with a valid Latin square and iteratively removes values.

\section{Case Studies}
\label{sec:case_study}
    \subsection{Evaluation Prompts}

        \begin{minipage}{0.48\linewidth}
            \begin{tcolorbox}[colback=white, colframe=gray!50!black, title={Sudoku}]
            Solve this Sudoku puzzle \textit{step-by-step} from top-left to bottom-right. Identify the empty cell and fill in the correct digit.
            \end{tcolorbox}
        \end{minipage}
            \hfill
        \begin{minipage}{0.48\linewidth}
            \begin{tcolorbox}[colback=white, colframe=gray!50!black, title={TSP}]
            Draw a continuous red line \textit{step-by-step} from the start point to form the shortest closed loop. Mark the current endpoint with a green dot while avoiding other circles.
            \end{tcolorbox}
        \end{minipage}
        
        
        \noindent
        \begin{minipage}{0.48\linewidth}
            \begin{tcolorbox}[colback=white, colframe=gray!50!black, title={VSP}]
            Draw a continuous red line from the Start point to the Goal point \textit{step-by-step}, avoiding holes. Mark the current end of the path with a green dot. Directions: D=Down, U=Up, R=Right, L=Left.
            \end{tcolorbox}
        \end{minipage}
            \hfill
        \begin{minipage}{0.48\linewidth}
            \begin{tcolorbox}[colback=white, colframe=gray!50!black, title={Maze}]
            Draw a continuous red line from the yellow dot to the blue dot \textit{step-by-step}, avoiding walls. Mark the current end of the path with a green dot. Directions: D=Down, U=Up, R=Right, L=Left.
            \end{tcolorbox}
        \end{minipage}

    \subsection{Qualitative Examples}

\begin{figure}[t]
    \centering
    \includegraphics[width=1\linewidth]{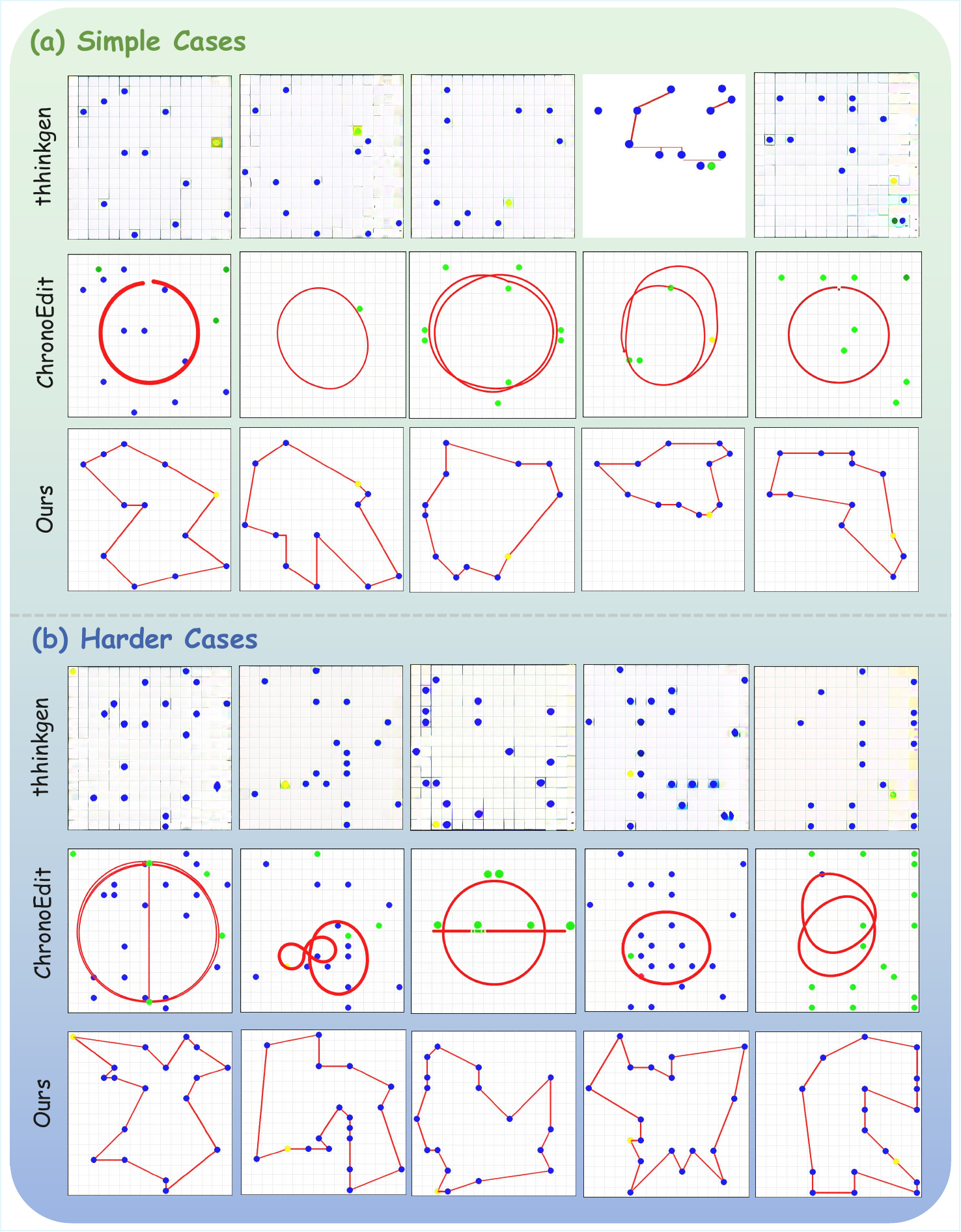}
    \caption{Qualitative comparison of EndoCoT and other methods on \textbf{TSP}.}
    \label{fig:more_case1}
\end{figure}

\begin{figure}[t]
    \centering
    \includegraphics[width=1\linewidth]{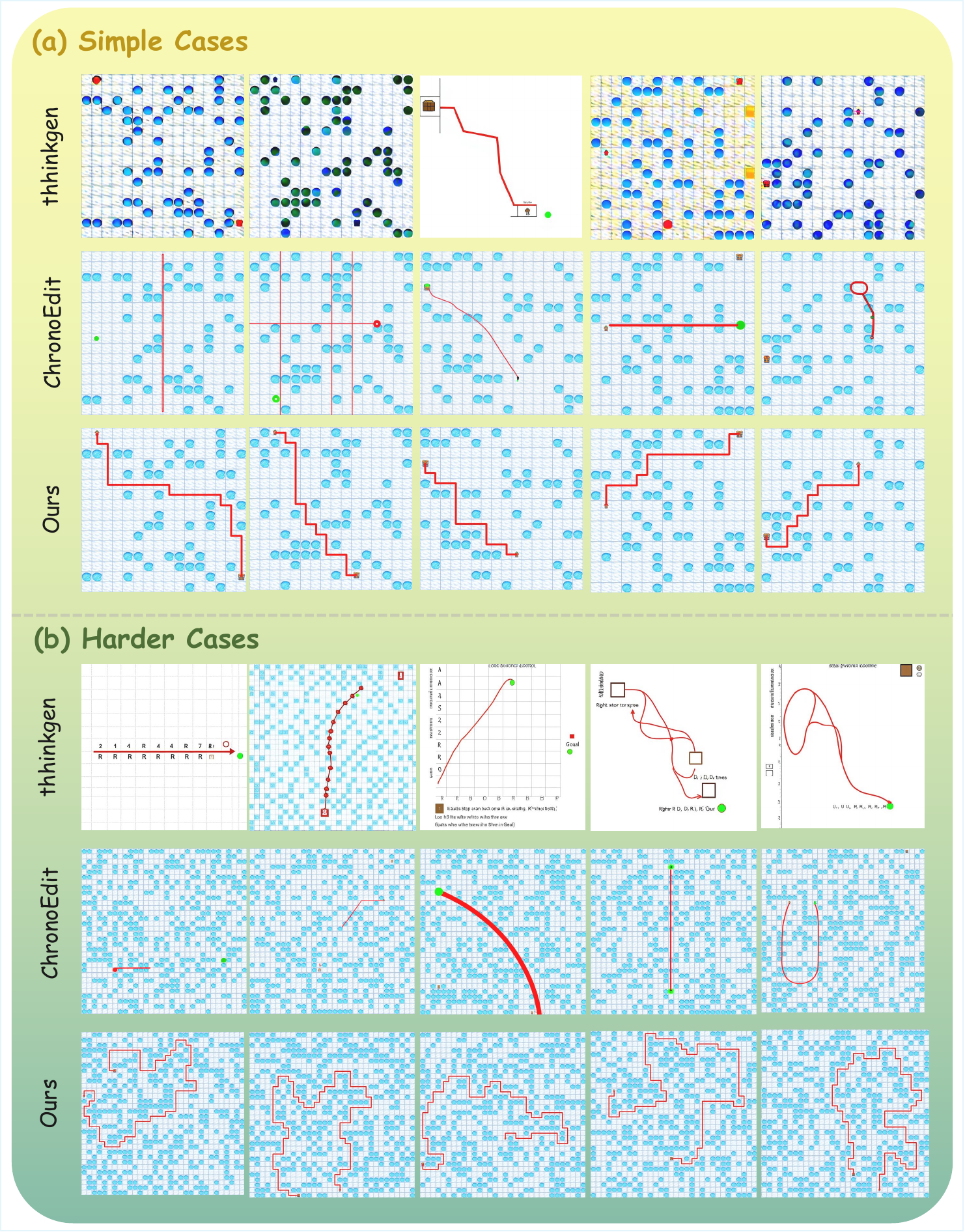}
    \caption{Qualitative comparison of EndoCoT and other methods on \textbf{VSP}.}
    \label{fig:more_case2}
\end{figure}

\begin{figure}[t]
    \centering
    \includegraphics[width=1\linewidth]{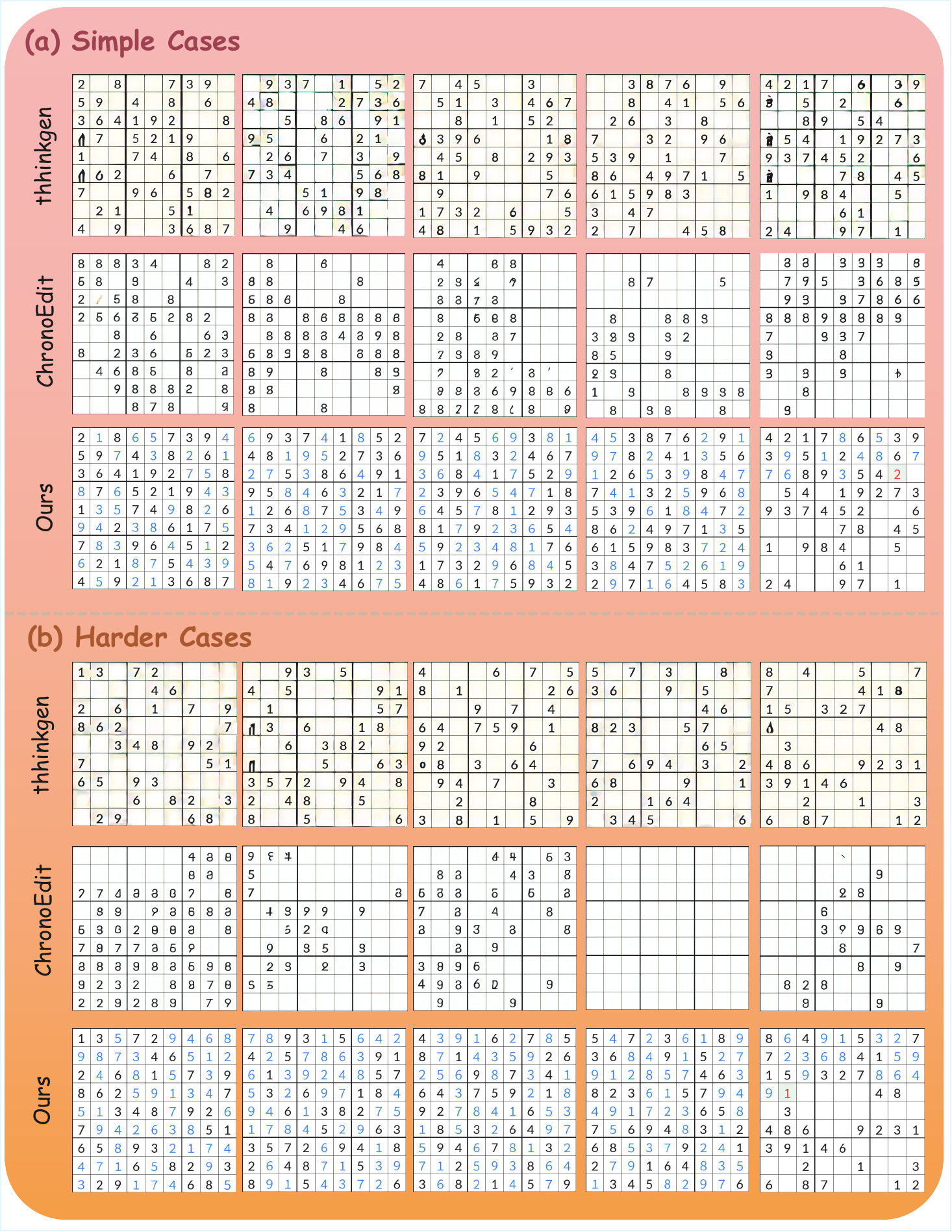}
    \caption{Qualitative comparison of EndoCoT and other methods on \textbf{Sudoku}.}
    \label{fig:more_case3}
\end{figure}

\begin{figure}[t]
    \centering
    \includegraphics[width=1\linewidth]{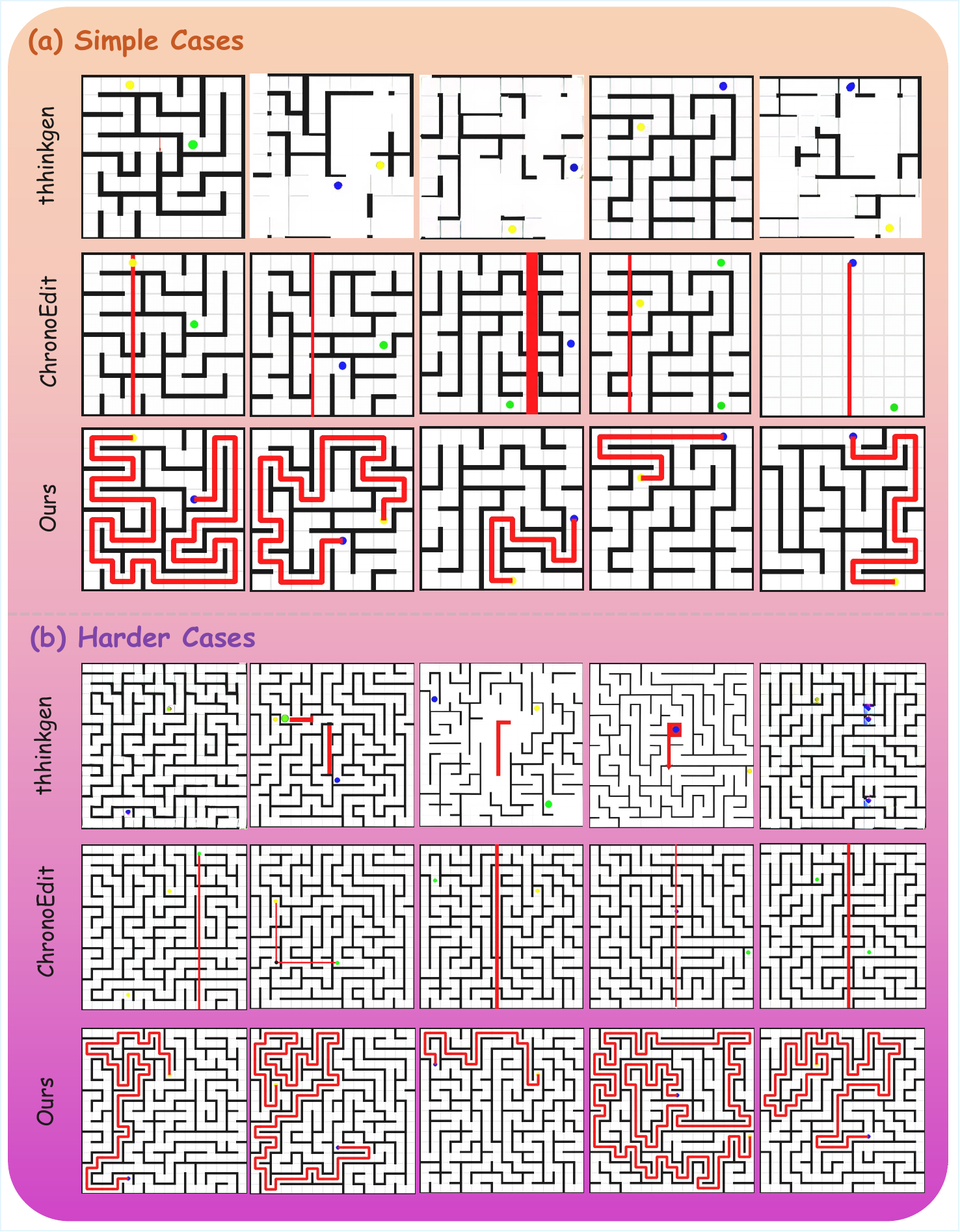}
    \caption{Qualitative comparison of EndoCoT and other methods on \textbf{Maze}.}
    \label{fig:more_case4}
\end{figure}

        
        Figs. \ref{fig:more_case1}, \ref{fig:more_case2}, \ref{fig:more_case3}, and \ref{fig:more_case4} provides additional qualitative examples comparing EndoCoT against ThinkGen \cite{jiao2025thinkgen} and ChronoEdit \cite{wu2025chronoedit} across various reasoning tasks.

\section{More Ablations}\label{app:more_ablations}

    \subsection{Effect of Two-Stage Training}

        \begin{table}[h]
            \centering 
            \caption{Single-stage training and multi-stage training on the Maze task.}
            \label{tab:training_settings}
            \begin{tabular}{l ccc ccc}
                \toprule
                \multirow{2}{*}{\textbf{Models}} & \multicolumn{3}{c}{\textbf{Accuracy (ACC)}} & \multicolumn{3}{c}{\textbf{Path Repetition (\%)}} \\
                \cmidrule(lr){2-4} \cmidrule(lr){5-7}
                & \textbf{8} & \textbf{16} & \textbf{32} & \textbf{8} & \textbf{16} & \textbf{32} \\
                \midrule
                Single-Stage  & 39 & 42 & 14 & 83.49 & 91.34 &  90.34 \\
                Two-Stage & \textbf{100} & \textbf{100} & \textbf{90} & \textbf{100.00} & \textbf{100.00} & \textbf{98.13} \\
                \bottomrule
        
            \end{tabular}
        \end{table}

        Tab.~\ref{tab:training_settings} presents an illustrative analysis on the Maze task, showing that Terminal Consolidation effectively enhances the model’s awareness of the final stage.
        
    \subsection{Impact of Terminal Consolidation Training Duration}

\begin{figure}[t]
          \centering
          
          \begin{minipage}{0.3\textwidth}
            \centering
            \includegraphics[width=\linewidth]{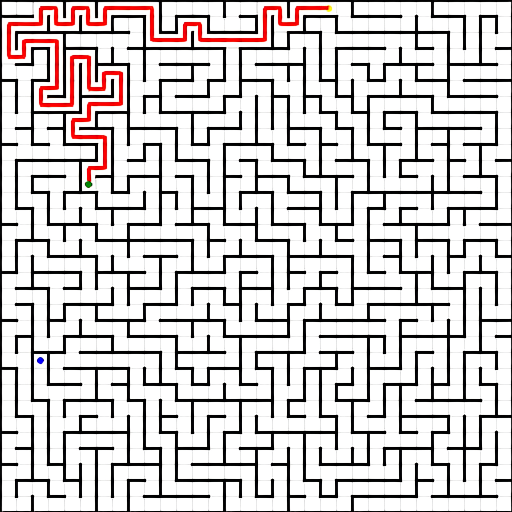}
            \vspace{-2em}
            \captionsetup{justification=centering}
            \caption*{steps=0, ${\tau=2}$}
          \end{minipage}
          \hfill
          \begin{minipage}{0.3\textwidth}
            \centering
            \includegraphics[width=\linewidth]{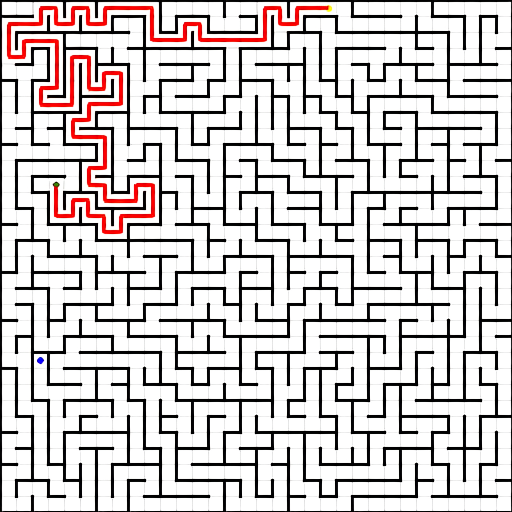}
            \vspace{-2em}
            \captionsetup{justification=centering}
            \caption*{steps=5000, ${\tau=2}$}
          \end{minipage}
          \hfill
          \begin{minipage}{0.3\textwidth}
            \centering
            \includegraphics[width=\linewidth]{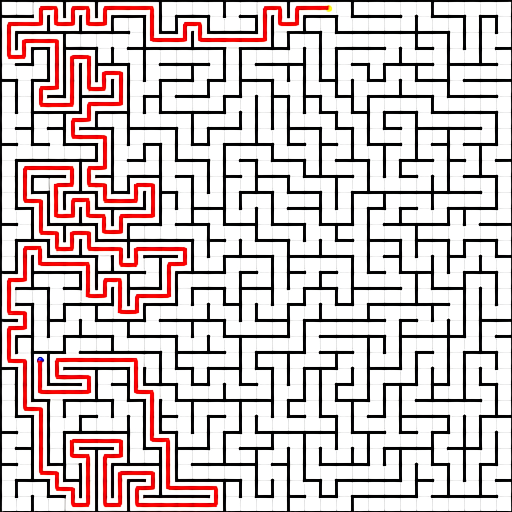}
            \vspace{-2em}
            \captionsetup{justification=centering}
            \caption*{steps=8000, ${\tau=2}$}
          \end{minipage}
        
          \vspace{1em}
        
          \begin{minipage}{0.3\textwidth}
            \centering
            \includegraphics[width=\linewidth]{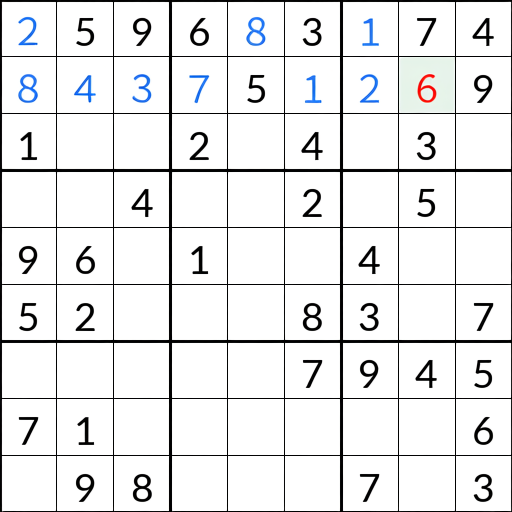}
            \vspace{-2em}
            \captionsetup{justification=centering}
            \caption*{steps=0, ${\tau=2}$}
          \end{minipage}
          \hfill
          \begin{minipage}{0.3\textwidth}
            \centering
            \includegraphics[width=\linewidth]{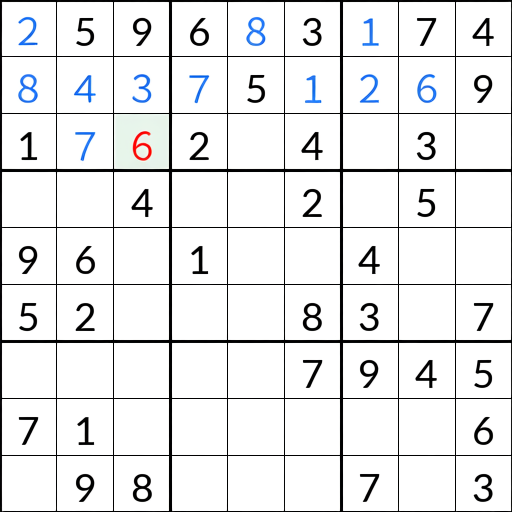}
            \vspace{-2em}
            \captionsetup{justification=centering}
            \caption*{steps=1000, ${\tau=2}$}
          \end{minipage}
          \hfill
          \begin{minipage}{0.3\textwidth}
            \centering
            \includegraphics[width=\linewidth]{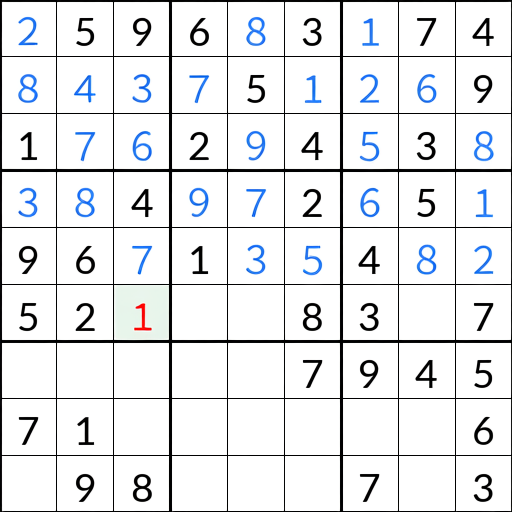}
            \vspace{-2em}
            \captionsetup{justification=centering}
            \caption*{steps=8000, ${\tau=2}$}
          \end{minipage}
        
          \caption{\textbf{Terminal Consolidation's Excessive Training Results.}}
          \label{fig:stage2_steps}
        \end{figure}

        Excessive training in Terminal Consolidation (Stage 2) causes intermediate steps to become sparse, as illustrated in Fig. \ref{fig:stage2_steps}. When choosing a constant number of reasoning steps $\tau$, checkpoints trained with more stage 2 training steps exhibit an inference behavior that converges more toward the target state.

    \subsection{Generalization under OOD Settings}

\begin{figure}[h]
            \centering
            \setlength{\tabcolsep}{1pt}
            
            \begin{tabular}{cc @{\hspace{8pt}} cc @{\hspace{8pt}} cc}
            
            \multicolumn{2}{c}{\small \hspace{-14pt}\textbf{Different Fonts}} &
            \multicolumn{2}{c}{\small \hspace{-12pt}\textbf{High Resolution}} &
            \multicolumn{2}{c}{\small \textbf{Low Resolution}} \\
            
            \addlinespace[2pt]
            
            \scriptsize Baseline & \scriptsize Ours &
            \scriptsize Baseline & \scriptsize Ours &
            \scriptsize Baseline & \scriptsize Ours \\
            
            \addlinespace[4pt]

            \includegraphics[width=0.15\textwidth]{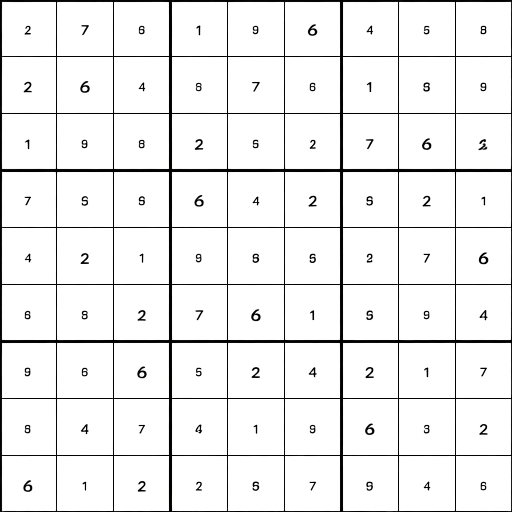} &
            \includegraphics[width=0.15\textwidth]{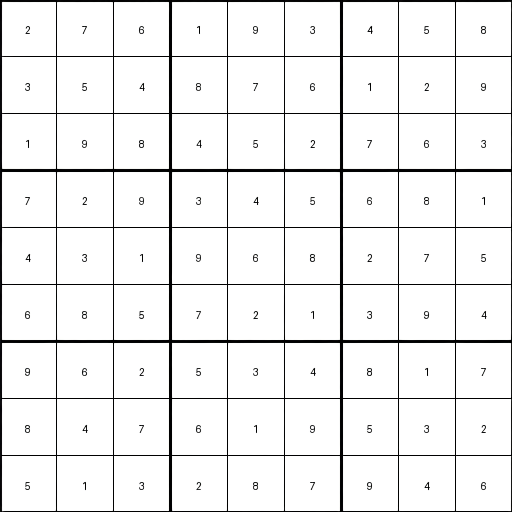} & 
            
            \includegraphics[width=0.15\textwidth]{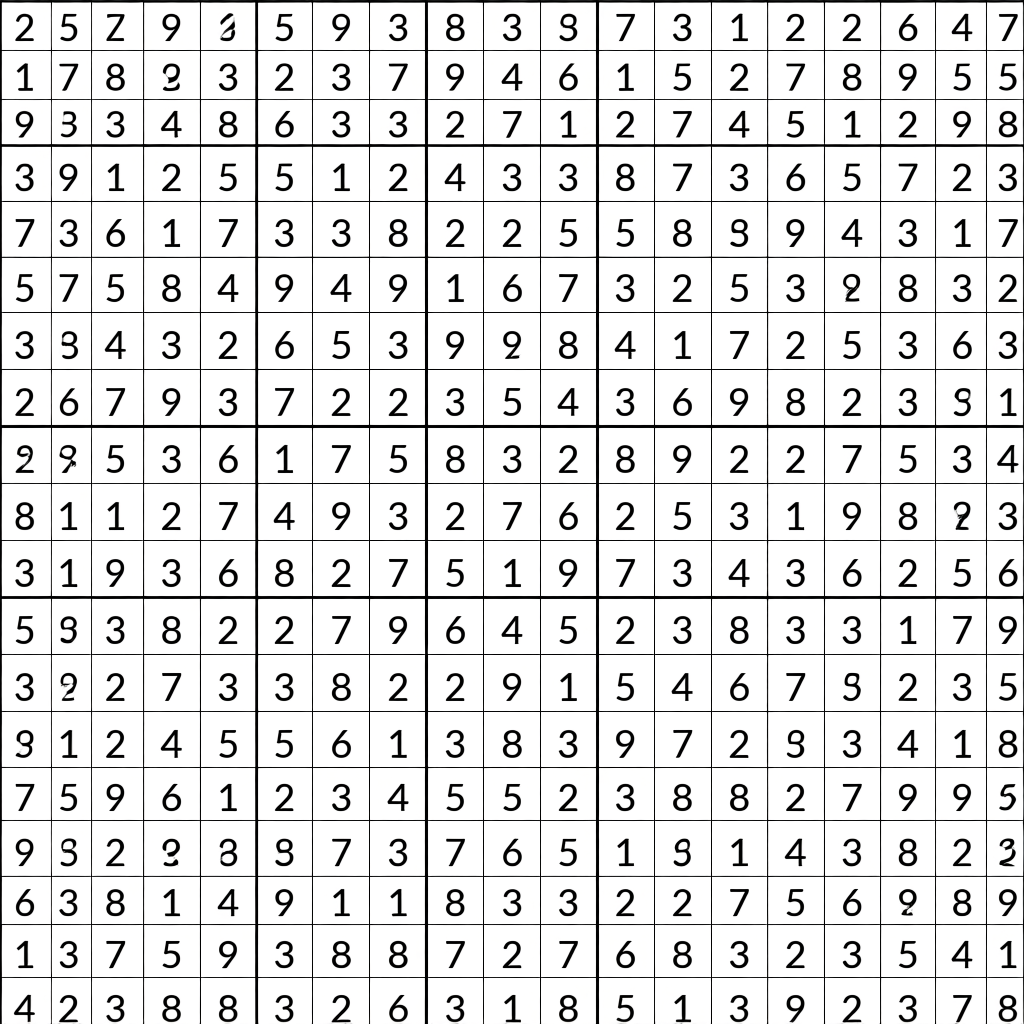} &
            \includegraphics[width=0.15\textwidth]{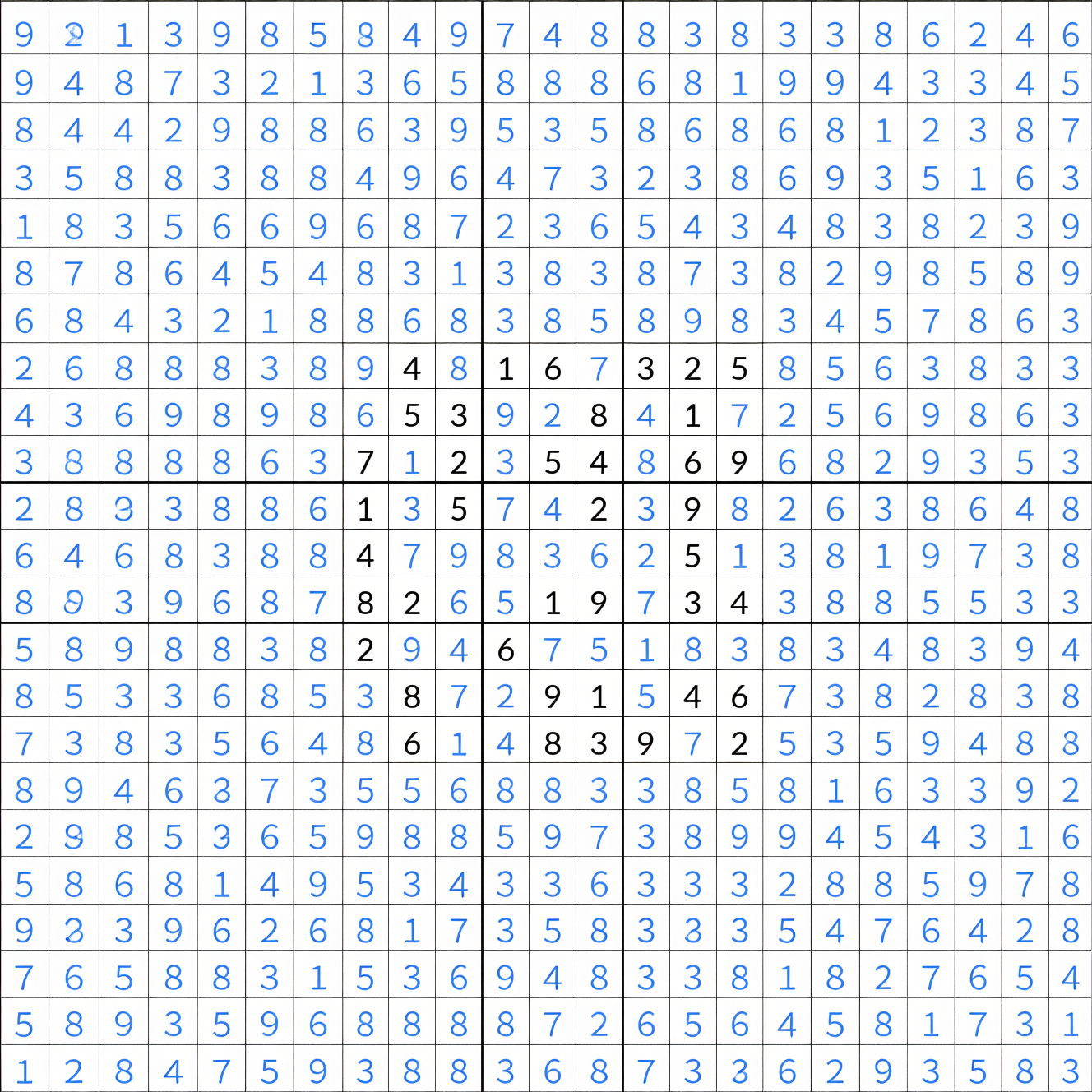} & 

            \includegraphics[width=0.15\textwidth]{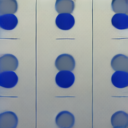} &
            \includegraphics[width=0.15\textwidth]{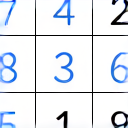} 
            
            \end{tabular}
            
            
            \caption{\small
            \textbf{OOD Test Result.} EndoCoT demonstrates stronger generalization than baseline methods when solving Sudoku puzzles with varying resolutions and font styles.
            }
            
            \label{fig:ood_comparison_row}
            
            
        \end{figure}

        On the Sudoku task, EndoCoT demonstrates stronger generalization than baseline methods across \textit{different resolutions} and \textit{font styles}, as shown in Fig.~\ref{fig:ood_comparison_row}.
        For high-resolution inputs, EndoCoT consistently produces correct predictions for the 9x9 grid located in the center. For low-resolution inputs, it can recover partially cropped correct answers, while DiffThinker fails to generate valid solutions.

\section{Training Hints}\label{app:training_hints}

\subsection{Training Settings}

    \begin{table}[H]
    \centering
    \caption{Training hyperparameters and LoRA configuration.}
    \label{tab:hyperparams}
    \begin{tabularx}{\textwidth}{@{} l l c l >{\raggedright\arraybackslash}X @{}}
        \toprule
        \multicolumn{2}{@{}l}{\textbf{General Settings}} & \phantom{a} & \multicolumn{2}{l@{}}{\textbf{LoRA Target Modules}} \\
        \cmidrule(r){1-2} \cmidrule(l){4-5}
        \textbf{Hyperparameter} & \textbf{Value} && \textbf{Component} & \textbf{Modules} \\
        \midrule
        Base Model & Qwen Image Edit 2511 && DiT Attn. & \texttt{to\_q/k/v}, \texttt{add\_*\_proj}, \texttt{to\_*\_out} \\
        Learning Rate & $1 \times 10^{-4}$ && DiT FFN & \texttt{img/txt\_mlp.net.2} \\
        Training Epochs & 5 && DiT Mod. & \texttt{img/txt\_mod.1} \\
        LoRA Rank & 32 && Text Enc. Attn. & \texttt{q/k/v/o\_proj} \\
        LoRA Target & DiT + Text Enc. && Text Enc. FFN & \texttt{gate/up/down\_proj} \\
    \bottomrule
    \end{tabularx}
    \end{table}

    The detailed training settings, including hyperparameters and LoRA configurations, are summarized in Tab.~\ref{tab:hyperparams}.

    \subsection{Further Analysis}
        \begin{figure}[t]
            \centering
            \includegraphics[width=1\linewidth]{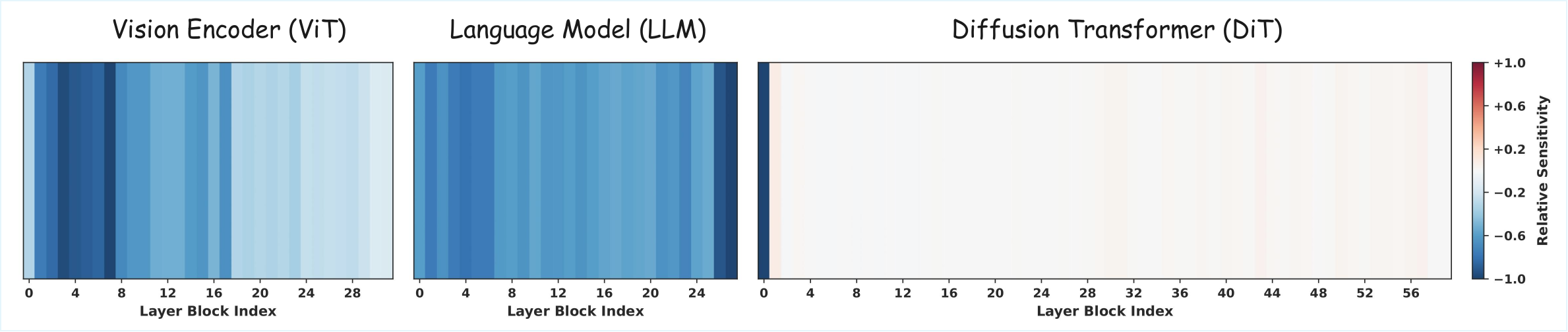}
            \caption{\textbf{Gradient difference between reasoning and non-reasoning datasets.}}
            \label{fig:further_analysis}
        \end{figure}

        To further investigate where reasoning-related signals are primarily encoded in the model, we analyze the gradient activation differences between datasets that require explicit reasoning and those that do not. Specifically, we compute the layer-wise gradient magnitude difference across the MLLM and DiT.
        
        As shown in Fig.~\ref{fig:further_analysis}, the largest gradient differences are consistently concentrated within the MLLM layers. This result further supports the analysis in Sec. 2: Reasoning capabilities are primarily mediated by the MLLM.

    \subsection{Experimental Observations}

    We obtain the following empirical observations while conducting experiments:
    \noindent \textbf{Fine-grained step supervision is crucial.}
    The number of reasoning steps performed during inference is typically much smaller than the number of steps supervised during training. Therefore, providing finer-grained intermediate step supervision is necessary for the model to  learn latent reasoning processes. \par
    \noindent \textbf{More semantic supervision is not necessarily better.}
    Applying semantic loss to intermediate steps can disrupt the synergy between the DiT rendering capability and the MLLM's understanding. Empirically, this leads to inferior performance compared to the sparse semantic supervision strategy in EndoCoT. \par
    \noindent \textbf{Number of latent tokens depend on algorithmic and visual complexity.}
    The number of latent tokens required to solve a puzzle is determined not only by the algorithmic complexity of the task, but also by its visual complexity. 

\section{Image Editing Details}\label{app:edit_data}
\begin{figure}[h]
    \centering
    \includegraphics[width=1\linewidth]{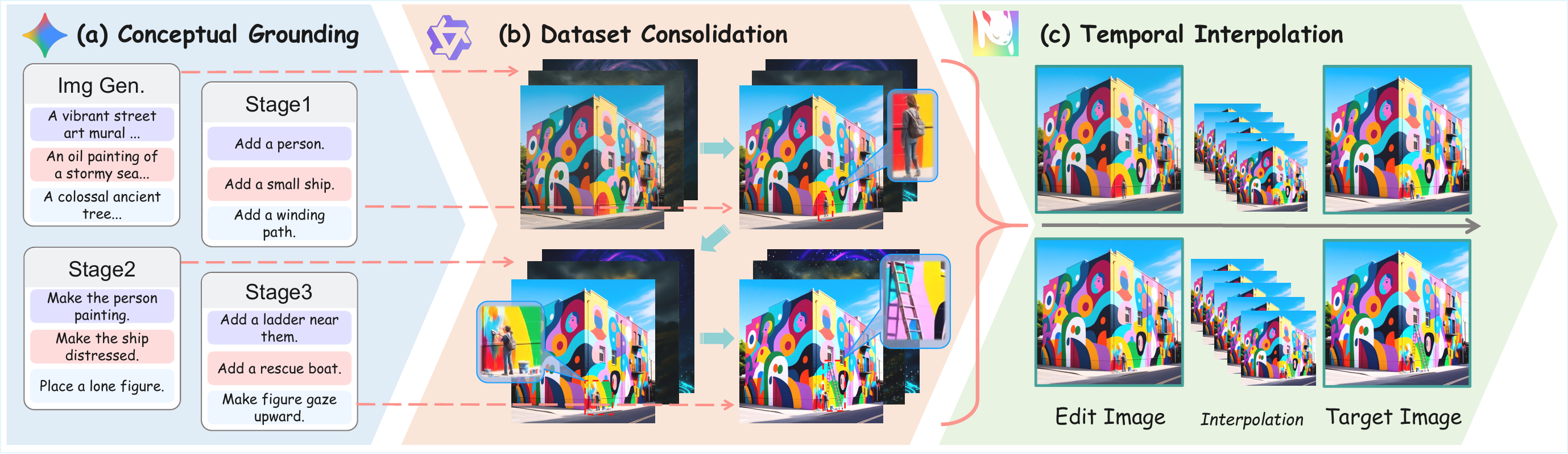}
    \caption{\textbf{Overview of the multi-step image editing dataset generation pipeline.}}
    \label{fig:edit_pipeline}
\end{figure}
We create a high-quality multi-step image editing dataset containing 10,000 unique scenes and 30,000 images.
To ensure diversity and quality, Gemini 2.5 \cite{google2026gemini3} first generates diverse initial scene descriptions paired with multi-step editing instructions.
Then Qwen-Image-Edit-2511 \cite{wu2025qwen} generate the corresponding intermediate and final edited images, providing strict per-step supervision. Ultimately, to ensure editing continuity, we optionally employ RIFE \cite{huang2022real} for intermediate frame interpolation.
The complete data generation pipeline is shown in Fig. \ref{fig:edit_pipeline}.

\begin{figure}[h]
    \centering
    
    \begin{minipage}[t]{0.48\textwidth}
        \begin{algorithm}[H]
            \caption{Maze Generation}
            \label{alg:maze}
            \begin{algorithmic}[1]
                \Require Grid dimension $N$
                \State Initialize grid graph $G = (\mathcal{V}, \mathcal{E}_{\emptyset})$ where $\mathcal{V} = \{1..N\}^2$
                \State $M \leftarrow \text{DFS\_SpanningTree}(G)$ \Comment{Perfect maze}
                \State Sample $s, g \sim \mathcal{U}(\mathcal{V})$ without replacement
                \State $P \leftarrow \text{BFS}(M, s, g)$ \Comment{Path $P = (v_1, \dots, v_K)$}
                \State $\mathcal{S} \leftarrow \emptyset$ \Comment{Intermediate visual states}
                \For{$k = 1$ \textbf{to} $K$}
                    \State $I_k \leftarrow \text{Render}(M, P_{1:k})$
                    \State $\mathcal{S} \leftarrow \mathcal{S} \cup \{I_k\}$
                \EndFor
                \State \Return Dataset instance $\mathcal{D} = (M, P, \mathcal{S})$
            \end{algorithmic}
        \end{algorithm}
    \end{minipage}
    \hfill
    \begin{minipage}[t]{0.48\textwidth}
        \begin{algorithm}[H]
            \caption{TSP Generation}
            \label{alg:tsp}
            \begin{algorithmic}[1]
                \Require 2D plane dimensions $W \times H$, city count $N_c$
                \State Sample cities $\mathbf{C} = \{c_1, \dots, c_{N_c}\} \sim \mathcal{U}(W \times H)$
                \State Compute distance matrix $\mathbf{D} \in \mathbb{R}^{N_c \times N_c}$ 
                \State $P \leftarrow \text{HeldKarpDP}(\mathbf{D})$ \Comment{Optimal tour}
                \State $\mathcal{S} \leftarrow \emptyset$
                \For{$k = 1$ \textbf{to} $N_c$}
                    \State $I_k \leftarrow \text{RenderTrajectory}(\mathbf{C}, P_{1:k})$
                    \State $\mathcal{S} \leftarrow \mathcal{S} \cup \{I_k\}$
                \EndFor
                \State \Return Dataset instance $\mathcal{D} = (\mathbf{C}, P, \mathcal{S})$
            \end{algorithmic}
        \end{algorithm}
    \end{minipage}
    
    

    \begin{minipage}{0.48\textwidth}
        \begin{algorithm}[H]
            \caption{VSP Generation}
            \label{alg:vsp}
            \begin{algorithmic}[1]
                \Require Map dimension $N$
                \State Generate map $\mathcal{M}$ with grid $\mathcal{V} = \{1..N\}^2$
                \State Assign types $\mathcal{T}(v) \in \{\text{Safe}, \text{Hole}\} \sim \text{Bern}(p_{hole})$
                \State Construct graph $\mathcal{G} = (\mathcal{V}_{safe}, \mathcal{E}_{safe})$
                \State Sample $s, g \sim \mathcal{U}(\mathcal{V}_{safe})$
                \State $P \leftarrow \text{Dijkstra}(\mathcal{G}, s, g)$ \Comment{Compute shortest path}
                \State \textbf{Assert} $P \neq \emptyset$ \Comment{Ensure a valid path exists} 
                \State $\mathcal{S} \leftarrow \emptyset$
                \For{$k = 1$ \textbf{to} $|P|$}
                    \State $I_k \leftarrow \text{RenderEnv}(\mathcal{M}, P_{1:k})$
                    \State $\mathcal{S} \leftarrow \mathcal{S} \cup \{I_k\}$
                \EndFor
                \State \Return Dataset instance $\mathcal{D} = (\mathcal{M}, P, \mathcal{S})$
            \end{algorithmic}
        \end{algorithm}
    \end{minipage}
    \hfill
    \begin{minipage}{0.48\textwidth}
        \begin{algorithm}[H]
            \caption{Sudoku Generation}
            \label{alg:sudoku}
            \begin{algorithmic}[1]
                \Require Target number of holes $H_{target}$
                \State $\mathbf{L} \leftarrow \text{GenerateLatinSquare}()$ \Comment{$9 \times 9$ solution}
                \State $\mathbf{M} \leftarrow \mathbf{L}$, $h \leftarrow 0$
                \While{$h < H_{target}$}
                    \State Sample $c \sim \text{NonEmptyCells}(\mathbf{M})$
                    \State $v \leftarrow \mathbf{M}[c]$
                    \State $\mathbf{M}[c] \leftarrow \emptyset$ \Comment{Dig hole}
                    \If{$\text{NumSolutions}(\mathbf{M}) > 1$}
                        \State $\mathbf{M}[c] \leftarrow v$ \Comment{Revert to maintain uniqueness}
                    \Else
                        \State $h \leftarrow h + 1$
                    \EndIf
                \EndWhile
                \State $\mathcal{S} \leftarrow \text{SimulateSolving}(\mathbf{M}, \mathbf{L})$
                \State \Return Dataset instance $\mathcal{D} = (\mathbf{M}, \mathbf{L}, \mathcal{S})$
            \end{algorithmic}
        \end{algorithm}
    \end{minipage}

    \caption{Pseudocode for Dataset Generation}
    \label{fig:pseudocode}
    
\end{figure}
    
\end{document}